\documentclass[letterpaper]{article} 
\usepackage[preprint]{aaai2027}  
\usepackage[hyphens]{url}  
\usepackage{graphicx} 
\usepackage{natbib}  
\usepackage{caption} 
\usepackage{amsmath, amssymb, amsthm}

\usepackage{booktabs}
\usepackage{makecell}
\usepackage{multirow}
\usepackage{array}
\usepackage{arydshln}

\usepackage{algorithm}
\usepackage[noend]{algorithmic}

\usepackage{subcaption}

\usepackage{mathtools}
\usepackage{pifont}
\usepackage{nccmath}
\usepackage{nicefrac}
\usepackage{microtype}

\newcommand{\methodshort}[1]{\textsc{SAFE-Merge}}
\newcommand{\methodlong}[1]{\textbf{S}parse Risk-\textbf{A}ware Masking and Low-Rank Recovery \textbf{F}or General-Knowledge Pr\textbf{E}servation}
\newcommand{\cmark}{{\ding{51}}}
\newcommand{\xmark}{{\ding{55}}}
\newcommand{\rot}[1]{\rotatebox[origin=l]{30}{#1}}

\renewcommand{\algorithmiccomment}[1]{\bgroup\hfill $\triangleright$ ~#1\egroup}
\newcommand{\notrianglecomment}[1]{\bgroup\hfill~#1\egroup}

\title{SAFE-Merge: Data-Free Continual Model Merging with General Knowledge Preservation}

\author{
Zihuan Qiu\textsuperscript{1$\dagger$},\;
Zhiyang Liao\textsuperscript{1$\dagger$},\;
Chiyuan He\textsuperscript{1},\;
Yi Xu\textsuperscript{2},\\
Fanman Meng\textsuperscript{1*},\;
Linfeng Xu\textsuperscript{1},\;
Qingbo Wu\textsuperscript{1},\;
Hongliang Li\textsuperscript{1}
}
\affiliations{
\textsuperscript{1}University of Electronic Science and Technology of China, Chengdu, China\\
\textsuperscript{2}Dalian University of Technology, Dalian, China
}

\def\eqref#1{Eq.~(\ref{#1})}

\def\1{\bm{1}}

\DeclareMathAlphabet{\mathsfit}{\encodingdefault}{\sfdefault}{m}{sl}
\SetMathAlphabet{\mathsfit}{bold}{\encodingdefault}{\sfdefault}{bx}{n}

\begin{document}
\renewcommand{\textsc}[1]{#1}

\maketitle
\begingroup
\renewcommand{\thefootnote}{\fnsymbol{footnote}}
\footnotetext[1]{Corresponding author.}
\footnotetext[2]{These authors contributed equally; Zihuan Qiu led the project.}
\endgroup

\begin{abstract}
Data-free continual model merging must incorporate a stream of specialized models while retaining both pretrained general knowledge and previously acquired tasks, without access to task data. Existing methods mainly merge task updates by suppressing interference among downstream tasks; while this protects previously acquired tasks, it overlooks the safety of the pretrained knowledge itself, whose erosion degrades generalization to held-out distributions and weakens the foundation for future task acquisition. We propose \methodshort{}, a simple data-free continual-merging framework that first decides which parameter updates are safe to retain, and then recovers the task information lost through masking. Specifically, to ensure safety, risk-aware sparse masking selects parameter updates that carry task-specific information while posing low risk to general knowledge. Masked low-rank recovery then compensates for the lost task information using only the same retained parameter updates, while leaving all masked-out parameters strictly unchanged. Finally, the combined update is fused into the backbone, incurring no additional inference cost. Across vision and language benchmarks, \methodshort{} consistently achieves the best H-score. On longer CLIP task sequences, it substantially improves H-score over NUFILT while also achieving the highest accuracy.
\end{abstract}

\section{Introduction}
Foundation models are increasingly adapted into collections of specialized models, each tuned for a particular dataset, domain, or user. Deploying every specialized checkpoint is expensive, while joint retraining is often impossible because the original task data are private, distributed, or no longer available. Model merging offers an attractive alternative: combine independently fine-tuned parameters directly, producing a single model without replaying their training data \citep{ilharco2023editing,yadav2023ties,tangFusionBenchComprehensiveBenchmark2024}. In realistic systems, however, models arrive over time rather than all at once. Data-free continual model merging (DFCMM) must therefore update one deployed backbone using only the incoming task model, the current merged model, and their shared pretrained initialization \citep{liu2023tangent,porrello2025a,tang2025merging}.

Continual merging is usually framed as a conflict among task updates. This view is incomplete. Every incoming task vector is added to a backbone whose pretrained parameters encode broad visual, linguistic, and cross-domain regularities. Aggressive task accumulation can damage this general knowledge even when earlier downstream tasks appear well preserved. Preserving accuracy on the merged tasks therefore does not guarantee that the model's broader pretrained capabilities remain intact. Such damage not only degrades performance on held-out distributions but also weakens the foundation for acquiring later tasks. Hence, a continual merger should preserve both previously encountered tasks and the transferable knowledge present before specialization.

The data-free setting makes this goal unusually difficult. Without examples, one cannot directly measure which parameter changes alter general behavior or which components are indispensable for the new task. The merger must infer both from parameters alone. Uniform averaging treats all parameter updates alike and accumulates interference \citep{izmailov2018averaging,ilharco2023editing}. Sign- or magnitude-based sparsification ignores the function represented by each direction \citep{yadav2023ties}. Orthogonal projection can protect selected subspaces, but removing all components in overlapping directions is overly rigid because useful task information and general knowledge need not be cleanly separable \citep{tang2025merging,chengwhoever,qiu2026nullspace}. Data-adaptive methods can estimate parameter importance, but require calibration samples and therefore leave the strictly data-free regime \citep{yang2023adamerging,tang2024merging,qiu2025mingle}.

\noindent These considerations lead to the central question of this work:
\begin{center}
\setlength{\fboxsep}{7pt}
\fbox{\parbox{0.88\columnwidth}{\centering\small\itshape
How can data-free continual merging learn new tasks while preserving prior tasks and pretrained knowledge?}}
\end{center}


We address this problem with \methodshort{}, a parameter-space method that first decides \emph{where} an incoming update can be safely retained through risk-aware masking, and then determines \emph{how} to recover the task utility lost during masking without modifying the masked-out region. Motivated by the finding that input-side singular directions of task vectors reflect task-data subspaces \citep{qiu2026nullspace}, we use parameter singular directions as data-free knowledge proxies. For each parameter update, \methodshort{} contrasts its energy along pretrained general directions with that along directions unique to the new task, retaining a low-risk quantile through a sparse binary mask.

Masking alone, however, improves safety at the cost of plasticity by discarding components important to the incoming task. We therefore learn a low-rank recovery term constrained to the retained parameter updates; those outside the mask remain exactly zero and are never modified. Its preservation objective suppresses responses along accumulated-task directions to limit forgetting, while its task-recovery objective matches the incoming task vector in its dominant subspace to restore new-task performance using only the retained updates. The masked update and recovery term are then folded into the backbone, requiring neither auxiliary data nor additional inference parameters. Across vision and language settings, \methodshort{} consistently achieves the best H-score and competitive backward transfer without increasing inference cost.

To summarize, our main contributions are as follows:
\begin{itemize}
\item We develop a data-free, parameter-wise risk criterion that contrasts general-subspace overlap with task-unique energy and selects a sparse set of parameter updates for safe knowledge acquisition.
\item We introduce masked low-rank recovery with preservation and task-recovery objectives, restoring useful information suppressed by sparse masking without adding inference parameters.
\item We evaluate \methodshort{} across vision and language benchmarks and conduct detailed ablations; it achieves the best H-score in every setting, demonstrating a stronger balance between continual-merging accuracy and held-out generalization.
\end{itemize}

\section{Related Work}

\paragraph{Model Merging.}
Research on model merging seeks a single checkpoint that combines capabilities learned by separately adapted models. Parameter averaging offers the simplest construction \citep{utans1996weight,shoemake1985animating} and is partly motivated by mode connectivity between aligned solutions \citep{entezari2021role,ainsworth2022git}. Task Arithmetic instead represents each adaptation as a displacement from a shared initialization and composes these task vectors algebraically \citep{ilharco2023editing}. In practice, independently learned displacements are rarely disentangled \citep{ortiz2023task}, so their direct addition can introduce sign conflicts, scale imbalance, and destructive interference. Later methods address these effects by selecting consistent parameter updates \citep{yadav2023ties}, reweighting task contributions \citep{yu2024language}, imposing structure during adaptation \citep{liu2023tangent,porrello2025a}, or estimating merge coefficients from calibration examples \citep{yang2023adamerging,tang2024merging,qiu2025mingle}. These advances primarily study a fixed collection of models; they do not by themselves resolve how knowledge should be protected as new checkpoints arrive sequentially.

Continual merging replaces one-shot composition with a recursive process in which only the current merged backbone and a newly arriving task model are available. Dataless and training-free formulations make this setting attractive when datasets cannot be retained or shared \citep{jin2023dataless,liu2023tangent,porrello2025a}. Recent approaches organize incoming updates through geometric constraints. Orthogonal-projection methods remove directions associated with earlier tasks \citep{tang2025merging,chengwhoever}, while \citet{qiu2026nullspace} uses task-vector singular structure to perform null-space filtering for stability and plasticity. Other methods introduce adaptive signals or auxiliary data to guide sequential integration \citep{qiu2025mingle}. Although these techniques target forgetting among merged tasks, the pretrained backbone is typically treated as a common reference rather than as a source of transferable knowledge that must itself be preserved. \methodshort{} addresses this complementary issue by explicitly distinguishing low-risk task updates from directions associated with pretrained general knowledge, and by recovering task utility only over the selected safe parameter updates.

\paragraph{Continual Learning.}
Conventional continual learning studies how sequential optimization erases earlier capabilities \citep{mccloskey1989catastrophic}. Representative solutions penalize changes to important parameters \citep{Kirkpatrick2016OvercomingCF,zenke2017continual,aljundi2018memory}, transfer predictions from an earlier model \citep{Hou2019LearningAU,Douillard2020PODNetPO}, maintain replay samples \citep{Rebuffi2016iCaRLIC,liu2021rmm}, or allocate task-dependent capacity \citep{lee2017overcoming,qiu2023ism,zhou2024expandable,yu2024boosting,huang2024class}. Checkpoint-based strategies have also been explored as a means of moderating forgetting \citep{Mirzadeh2020LinearMC,wen2023optimizing,marczak2024magmax}. Most of these approaches assume access to training trajectories, stored examples, or expandable modules, assumptions unavailable in strict data-free merging. Our setting instead requires all preservation and acquisition decisions to be inferred from the pretrained, merged, and incoming parameters, with no task data and no persistent increase in model size.

\section{Motivation}
\subsection{Not All Parameters Are Equally Safe to Merge}

Let $\theta_0$ be a pretrained model and
$\tau_t=\theta_t-\theta_0$ the task vector of the model fine-tuned on task
$t$. We start from the premise that entries of $\tau_t$ overwrite the general
knowledge encoded in $\theta_0$ to varying degrees. Below, we derive a
data-free score for this overwrite risk and verify that its ordering indeed
identifies a safe parameter subset.

\textbf{Parameter risk from a second-order loss change.}
Let $\mathcal L_g(\theta)$ measure how well the general knowledge of the
pretrained model is preserved. Changing a single parameter
$\theta_j$ by $\delta_j$ changes this loss, to second order, by
\begin{equation}
\Delta\mathcal L_{g,j}
=\left[\nabla\mathcal L_g(\theta_0)\right]_j\delta_j
+\tfrac{1}{2}\left[\nabla^2\mathcal L_g(\theta_0)\right]_{jj}\delta_j^2
+o(\delta_j^2).
\label{eq:taylor_risk}
\end{equation}
Because $\theta_0$ is optimized for the pretraining objective, the first-order
term vanishes and the change is governed by curvature: a perturbation of a
given magnitude is cheap in a flat direction but costly in a sharp one. The
risk of modifying a parameter is thus determined by the \emph{curvature} of
$\mathcal L_g$ along that parameter.

\textbf{A computable surrogate for the curvature.}
We approximate the local preservation cost by the squared change in a linear
layer's response to general-domain inputs. Since
$(W_0^{(l)}+\Delta W^{(l)})x-W_0^{(l)}x=\Delta W^{(l)}x$, this proxy is
\begin{equation}
\begin{aligned}
\mathcal R_g^{(l)}(\Delta W)
&=\frac{1}{2}\mathbb E_{x\sim\mathcal D_g}
\left\|\Delta W^{(l)}x\right\|_2^2\\
&=\frac{1}{2}\operatorname{tr}\!\left(
\Delta W^{(l)}C_g^{(l)}
\Delta W^{(l)\top}\right).
\end{aligned}
\label{eq:data_curvature}
\end{equation}
Here $C_g^{(l)}=\mathbb E[xx^\top]$ is the uncentered activation covariance.
The proxy is zero and stationary at $\Delta W^{(l)}=0$, while $C_g^{(l)}$
plays the role of its curvature: a perturbation is costly in directions where
$C_g^{(l)}$ carries large energy.

\textbf{Estimating the sensitive directions without data.}
In data-free merging, $C_g^{(l)}$ is unavailable. Motivated by the observed
correspondence between representation directions and the input-side singular
directions of parameter updates \citep{qiu2026nullspace}, we approximate its
dominant subspace from the pretrained weight itself: treating
$\theta_0^{(l)}$ as a generalized update from the origin, we take its top-$r_p$
right singular vectors $V_0^{(l)}$ and define
\begin{equation}
Q_0^{(l)}=\operatorname{orth}(V_0^{(l)}),\qquad
\widehat C_g^{(l)}
=Q_0^{(l)}Q_0^{(l)\top}.
\label{eq:general_curvature}
\end{equation}
where $\operatorname{orth}(\cdot)$ returns an orthonormal basis for the column
space of its argument. Note that this uses the pretrained weight only to recover candidate directions
associated with general knowledge; it does not assume that pretraining started
from zero. Substituting $\Delta W^{(l)}=\tau_t^{(l)}$ and
$\widehat C_g^{(l)}$ into Eq.~(\ref{eq:data_curvature}) and distributing the
projected energy across entries then gives
\begin{equation}
R_{g,ij}^{(l)}
=
\left(\tau_t^{(l)}Q_0^{(l)}Q_0^{(l)\top}\right)_{ij}^2
.
\label{eq:general_risk}
\end{equation}
An entry scores high when its change aligns with the directions of the
pretrained model, and is therefore likely to overwrite general knowledge.
Keeping the lowest-scoring entries and zeroing out the rest yields a mask that
admits the update only where it is safe.

\textbf{Validating the general-risk score.}
We test whether the score in Eq.~(\ref{eq:general_risk}) correctly identifies
which parameter updates damage pretrained general knowledge. For each keep
ratio $\rho$, we rank the entries of an incoming task vector by
$R_{g,ij}^{(l)}$, retain the bottom-$\rho$ fraction, and reset the remaining
entries to their pretrained values. As a control, we reverse the ranking and
retain the top-$\rho$ fraction instead. We then evaluate on ImageNet,
ImageNet-R, and ObjectNet as held-out probes of general visual knowledge.

\begin{figure}[t]
\centering
\includegraphics[width=.9\columnwidth]{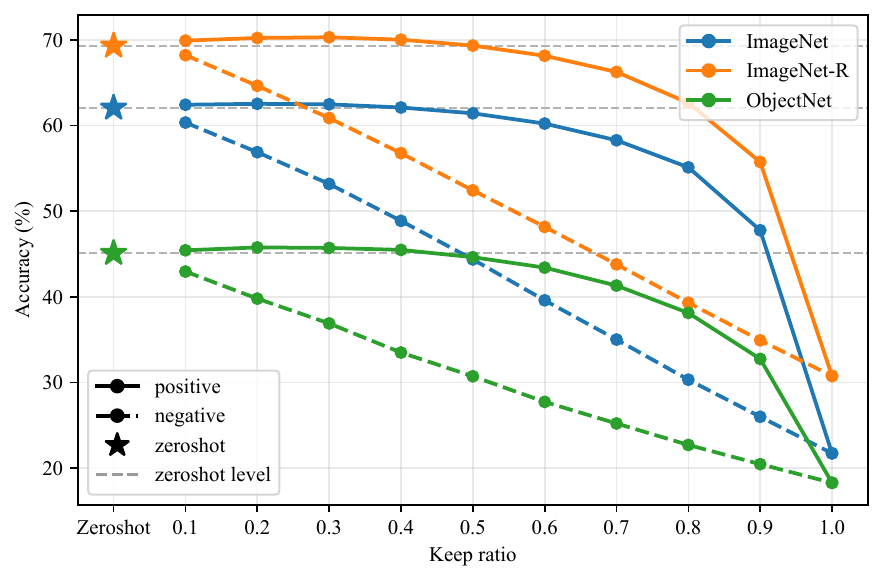}
\caption{\textbf{Risk score protects general knowledge during continual
merging.} General-domain accuracy on ViT-B/32 after sequentially merging 8
tasks. At each step, we retain a fraction $\rho$ of the incoming task vector
selected by the general-risk score in Eq.~(\ref{eq:general_risk}), apply the
resulting mask directly, and add the masked update to the running merged model.
\emph{Positive selection} (solid) retains low-risk parameter updates; \emph{negative
selection} (dashed) retains high-risk parameter updates. Both use exactly the same
parameter budget. Stars and horizontal dashed lines mark pretrained zero-shot
performance.}
\label{fig:risk_ratio}
\end{figure}

Fig.~\ref{fig:risk_ratio} confirms the score. We merge 8 tasks sequentially by
adding only the masked portion of each incoming task vector to the running
merged model. Retaining the lowest-risk parameter updates produces a broad flat region:
performance stays close to the pretrained model on all three datasets even when
up to $50\%$ of each task vector is kept, while reversing the order causes a
large drop under the same budget. Since the two selections differ only in which
end of the ranking they keep, this gap validates the risk score rather than
sparsity alone. As $\rho$ approaches $1$, both converge to the full task vector,
which substantially damages general performance. Thus, low-scored entries are
genuinely safer, whereas high-scored entries carry greater overwrite risk.

\section{Method: \methodshort{}}

\methodshort{} (\methodlong{}) addresses data-free continual merging in three steps:
\ding{182} compute a binary mask $M_t$ that admits parameter updates that are safe for
general knowledge and useful for the incoming task; \ding{183} learn a low-rank
recovery over the parameter updates retained by the mask to restore task information suppressed by
masking; and \ding{184} fold the masked update and the low-rank recovery term into the backbone.
We write $\theta_t^{\mathrm{merged}}$ for the merged backbone after task $t$ is
incorporated, with $\theta_0^{\mathrm{merged}}=\theta_0$.
No task data are used and no extra parameters remain at inference time.

\begin{figure*}[t]
\centering
\includegraphics[width=1\textwidth]{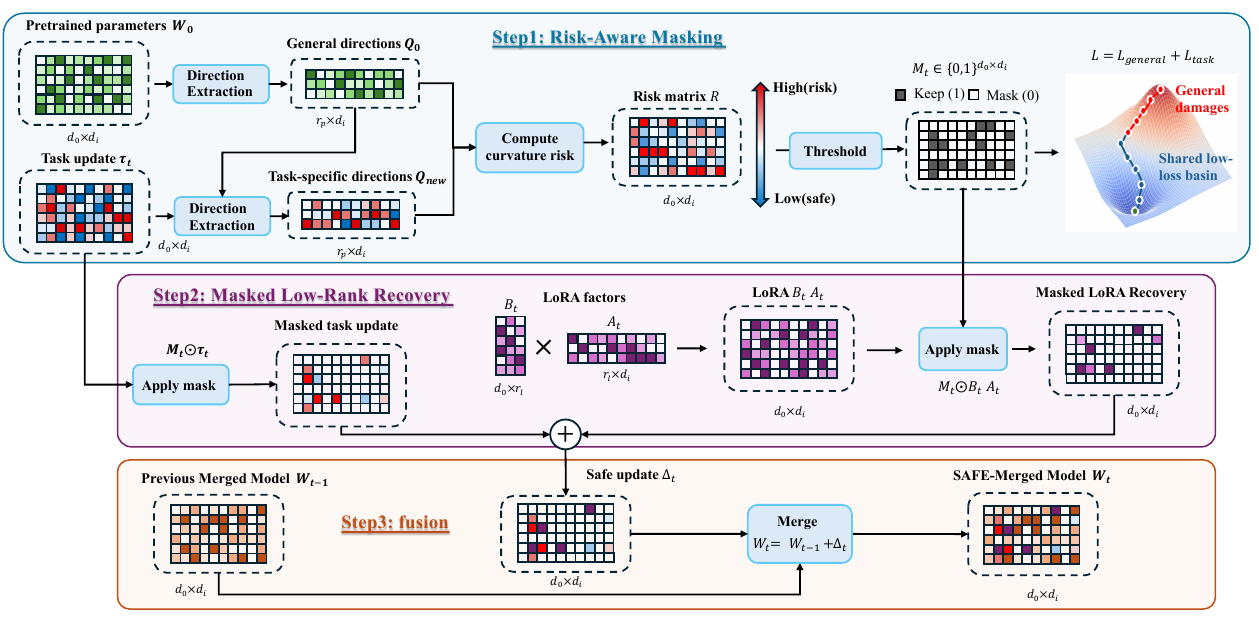}
\caption{\textbf{Overview of \methodshort{}.}
\methodshort{} identifies low-risk parameter updates in the incoming task vector, recovers
task information over the retained subset using a masked low-rank update, and
fuses the resulting update into the running merged model without retaining extra
inference-time parameters.}
\label{fig:safe_merge_method}
\end{figure*}

\subsection{Step 1: Risk-Aware Masking}
\label{sec:risk_routing}

For layer $l$, recall the incoming task vector $\tau_t^{(l)}=\theta_t^{(l)}-\theta_0^{(l)}$. Let $V_0^{(l)}$ and $V_t^{(l)}$ denote the top-$r_p$ right singular vectors of $\theta_0^{(l)}$ and $\tau_t^{(l)}$, respectively. To separate directions shared with the pretrained model from those unique to the incoming task, we orthogonalize $V_t^{(l)}$ against $V_0^{(l)}$:
\begin{equation}
\begin{split}
Q_0^{(l)}&=\operatorname{orth}(V_0^{(l)}),\\
Q_{\mathrm{new}}^{(l)}
&=\operatorname{orth}\!\left[(I-Q_0^{(l)}Q_0^{(l)\top})V_t^{(l)}\right].
\end{split}
\label{eq:risk_bases}
\end{equation}
The projector $Q_0^{(l)}Q_0^{(l)\top}$ captures directions associated with general knowledge, while $Q_{\mathrm{new}}^{(l)}Q_{\mathrm{new}}^{(l)\top}$ captures task-distinctive directions. The score in Eq.~(\ref{eq:general_risk}) penalizes only overwrite risk; here we additionally reward task utility, so that a parameter update that is risky yet vital for the incoming task is not masked out. The per-entry risk score thus contrasts preservation cost against task utility:
\begin{equation}
R_{ij}^{(l)}
=
\left(\tau_t^{(l)}Q_0^{(l)}Q_0^{(l)\top}\right)_{ij}^2
-
\left(\tau_t^{(l)}Q_{\mathrm{new}}^{(l)}Q_{\mathrm{new}}^{(l)\top}\right)_{ij}^2
.
\label{eq:risk_score}
\end{equation}
Large projected energy along $Q_0^{(l)}$ indicates high overwrite risk, while energy along $Q_{\mathrm{new}}^{(l)}$ indicates task-specific value; a parameter update is masked only when the former outweighs the latter.

Given a keep ratio $\rho\in[0,1]$, the mask retains the lowest-risk parameter updates:
\begin{equation}
M_t^{(l)}=\mathbb{I}\!\left[R_{ij}^{(l)}
\leq \operatorname{Quantile}_{\rho}(R_{ij}^{(l)})\right],
\label{eq:risk_mask}
\end{equation}
where $\mathbb{I}[\cdot]$ is the indicator function and
$\operatorname{Quantile}_{\rho}(\cdot)$ returns the $\rho$-th quantile over
all parameter updates of the layer.

\subsection{Step 2: Masked Low-Rank Recovery}

Masking improves safety by zeroing out high-risk coordinates, yet the discarded portion of $\tau_t^{(l)}$ may contain information needed for the incoming task. We restore this information with a low-rank recovery term of rank $r_l$, constrained to the same retained parameter updates:
\begin{equation}
\Delta_t^{(l)}
=M_t^{(l)}\odot\tau_t^{(l)}
+M_t^{(l)}\odot\left(B_t^{(l)}A_t^{(l)}\right),
\label{eq:recovered_update}
\end{equation}
where $\odot$ denotes the element-wise product, $A_t^{(l)}\in\mathbb{R}^{r_l\times d_i}$ and $B_t^{(l)}\in\mathbb{R}^{d_o\times r_l}$. Note that the mask is applied \emph{after} forming the product $B_t^{(l)}A_t^{(l)}$, so the recovery term cannot modify any coordinate that masking rejected as high-risk: every masked-out parameter update remains exactly zero.

The recovery factors are optimized without task data by projecting the update onto parameter-derived task subspaces, following the same direction-based proxy as \citet{qiu2026nullspace}. Specifically, the directions used by earlier tasks are estimated from the accumulated merged vector $\bar\tau_{t-1}^{(l)}=\theta_{t-1}^{\mathrm{merged},(l)}-\theta_0^{(l)}$, and those of the incoming task from $\tau_t^{(l)}$ itself: we take the top-$r_p$ right singular vectors $V_{\mathrm{old}}^{(l)}$ of $\bar\tau_{t-1}^{(l)}$ and the top-$r_v$ right singular vectors $V_{\mathrm{new}}^{(l)}$ of $\tau_t^{(l)}$, and define the projectors
\begin{equation}
G_{\mathrm{old}}^{(l)}
=V_{\mathrm{old}}^{(l)}V_{\mathrm{old}}^{(l)\top},
\qquad
G_{\mathrm{new}}^{(l)}
=V_{\mathrm{new}}^{(l)}V_{\mathrm{new}}^{(l)\top}.
\end{equation}

The two recovery objectives act on complementary subspaces. To limit forgetting, we penalize the component of the recovered update that lies along previously accumulated task directions:
\begin{equation}
\mathcal L_{\mathrm{old}}^{(l)}
=\left\|\Delta_t^{(l)}G_{\mathrm{old}}^{(l)}\right\|_F^2.
\label{eq:old_loss}
\end{equation}
To restore plasticity, we match the merged update to the task vector on the new task's dominant subspace:
\begin{equation}
\mathcal L_{\mathrm{new}}^{(l)}
=\left\|\tau_t^{(l)}G_{\mathrm{new}}^{(l)}
-\left(\bar\tau_{t-1}^{(l)}+\Delta_t^{(l)}\right)G_{\mathrm{new}}^{(l)}\right\|_F^2.
\label{eq:new_loss}
\end{equation}
Intuitively, $\mathcal L_{\mathrm{old}}^{(l)}$ pushes the update away from the subspaces of earlier tasks, while $\mathcal L_{\mathrm{new}}^{(l)}$ pulls it toward the incoming task's behavior on its own subspace. The full layer-wise objective combines these terms with a regularizer on the recovery magnitude:
\begin{equation}
\mathcal L^{(l)}
=\lambda\mathcal L_{\mathrm{new}}^{(l)}
+(1-\lambda)\mathcal L_{\mathrm{old}}^{(l)}
+\mu\left\|M_t^{(l)}\odot B_t^{(l)}A_t^{(l)}\right\|_F^2,
\label{eq:stacked_proj_loss}
\end{equation}
where $\lambda\in[0,1]$ trades plasticity for stability and $\mu$ regularizes the recovery term. We optimize $\sum_l \mathcal L^{(l)}$ over selected layers by gradient descent without accessing any task data.

\subsection{Step 3: Fusion}

After optimizing the recovery factors, the final update is added to the previous merged weights:
\begin{equation}
\label{eq:merge}
\theta_t^{\text{merged},(l)}
= \theta_{t-1}^{\text{merged},(l)}
+\Delta_t^{(l)}.
\end{equation}
The mask and low-rank factors are discarded after fusion, leaving the parameter count and inference cost unchanged. The full procedure is summarized in Alg.~\ref{alg:mainalgo}.

\begin{algorithm}[t]
\small
\caption{\methodshort{} Procedure}
\label{alg:mainalgo}
\begin{algorithmic}
  \STATE {\bfseries Input:} $\theta_0$, $\{\theta_t\}_{t=1}^T$, selected layers $\{1,\dots,L\}$, ranks $\{r_p,r_l,r_v\}$, keep ratio $\rho$, learning rate $\eta$, iterations $K$, weights $\lambda,\mu$.
  \STATE \textbf{Initialize:} $\theta_0^{\text{merged}} \leftarrow \theta_0$
  \vspace{0.25em}
  \FOR{each task $t = 1,\dots,T$}
    \FOR{each selected layer $l$}
      \STATE $\tau_t^{(l)} \leftarrow \theta_t^{(l)} - \theta_0^{(l)}$;\quad $\bar\tau_{t-1}^{(l)} \leftarrow \theta_{t-1}^{\text{merged},(l)} - \theta_0^{(l)}$
      \STATE Build $Q_0^{(l)}$, $Q_{\mathrm{new}}^{(l)}$ via Eq.~(\ref{eq:risk_bases})
      \STATE Compute $R_{ij}^{(l)}$ via Eq.~(\ref{eq:risk_score}) and $M_t^{(l)}$ via Eq.~(\ref{eq:risk_mask})
      \STATE Initialize $A_t^{(l)}$ randomly, $B_t^{(l)} \leftarrow \mathbf{0}$
    \ENDFOR
    \FOR{$k = 1,\dots,K$}
      \FOR{each selected layer $l$}
        \STATE Build $G_{\mathrm{old}}^{(l)}$ from $\bar\tau_{t-1}^{(l)}$ (zero if $t=1$)
        \STATE Build $G_{\mathrm{new}}^{(l)}$ from $\tau_t^{(l)}$
        \STATE $\Delta_t^{(l)} \leftarrow M_t^{(l)} \odot \big(\tau_t^{(l)} + B_t^{(l)}A_t^{(l)}\big)$
        \STATE Update $A_t^{(l)}, B_t^{(l)}$ by gradient descent on $\mathcal{L}^{(l)}$ (Eq.~(\ref{eq:stacked_proj_loss}))
      \ENDFOR
    \ENDFOR
    \FOR{each selected layer $l$}
      \STATE $\theta_t^{\text{merged},(l)} \leftarrow \theta_{t-1}^{\text{merged},(l)} + M_t^{(l)} \odot \big(\tau_t^{(l)} + B_t^{(l)}A_t^{(l)}\big)$
    \ENDFOR
  \ENDFOR
  \vspace{0.25em}
  \STATE \textbf{Output:} $\theta_T^{\text{merged}}$
\end{algorithmic}
\end{algorithm}

\section{Experiments}

\begin{table*}[t]
\centering
\caption{
Results on ViT-B/32 over ten task orders. We report continual merging accuracy
(ACC) and backward transfer (BWT) on the merged tasks, general performance
(Gen.) on held-out control datasets, and their harmonic mean (H-score).
Best results are in \textbf{bold} and second-best are \underline{underlined}.
}
\setlength{\tabcolsep}{3.2pt}
\renewcommand\arraystretch{1.18}
\newcommand{\pmv}[2]{#1{\scriptsize\textsubscript{\ensuremath{\pm#2}}}}
\resizebox{\textwidth}{!}{%
\begin{tabular}{l|cccc|cccc|cccc}
\toprule
\multirow{2}{*}{\textbf{Method}}
& \multicolumn{4}{c|}{\textbf{8 Tasks}}
& \multicolumn{4}{c|}{\textbf{14 Tasks}}
& \multicolumn{4}{c}{\textbf{20 Tasks}} \\
\cmidrule(lr){2-5}
\cmidrule(lr){6-9}
\cmidrule(lr){10-13}
& \textbf{ACC$\uparrow$} & \textbf{Gen.$\uparrow$} & \textbf{H-score$\uparrow$} & \textbf{BWT$\uparrow$}
& \textbf{ACC$\uparrow$} & \textbf{Gen.$\uparrow$} & \textbf{H-score$\uparrow$} & \textbf{BWT$\uparrow$}
& \textbf{ACC$\uparrow$} & \textbf{Gen.$\uparrow$} & \textbf{H-score$\uparrow$} & \textbf{BWT$\uparrow$} \\
\midrule

\textsc{Pre-Trained}
& 48.1 & 58.8 & 52.9 & --
& 56.9 & 58.8 & 57.8 & --
& 55.6 & 58.8 & 57.1 & -- \\

\textsc{Individual}
& 90.4 & 42.5 & 57.8 & --
& 89.3 & 44.4 & 59.3 & --
& 89.8 & 44.4 & 59.4 & -- \\

\midrule

\textsc{Task Arithmetic}
& \pmv{67.5}{0.0} & \pmv{44.5}{0.0} & \pmv{53.6}{0.0} & \pmv{-9.6}{1.5}
& \pmv{66.5}{0.0} & \pmv{57.3}{0.0} & \pmv{61.6}{0.0} & \pmv{\underline{-1.3}}{1.6}
& \pmv{60.0}{0.0} & \pmv{\underline{53.8}}{0.0} & \pmv{56.7}{0.0} & \pmv{-3.4}{1.0} \\

\textsc{Ties-Merging}
& \pmv{49.0}{10.2} & \pmv{21.3}{7.7} & \pmv{29.5}{8.8} & \pmv{-15.3}{8.0}
& \pmv{66.2}{0.6} & \pmv{\underline{57.8}}{0.3} & \pmv{61.8}{0.4} & \pmv{\textbf{1.9}}{0.6}
& \pmv{59.9}{0.7} & \pmv{51.4}{1.3} & \pmv{55.2}{1.0} & \pmv{\underline{-1.5}}{0.7} \\

\textsc{WUDI-Merging}
& \pmv{74.7}{6.6} & \pmv{43.5}{7.3} & \pmv{54.9}{7.5} & \pmv{-17.0}{7.5}
& \pmv{67.0}{6.9} & \pmv{43.7}{5.7} & \pmv{52.8}{6.2} & \pmv{-22.8}{7.3}
& \pmv{63.7}{3.8} & \pmv{45.0}{4.1} & \pmv{52.2}{3.9} & \pmv{-26.0}{4.1} \\

\textsc{Iso-C}
& \pmv{71.7}{1.2} & \pmv{\underline{55.5}}{0.6} & \pmv{62.9}{0.5} & \pmv{-10.2}{1.2}
& \pmv{73.2}{1.8} & \pmv{53.4}{0.6} & \pmv{61.6}{0.7} & \pmv{-10.4}{1.9}
& \pmv{67.6}{0.8} & \pmv{48.8}{0.9} & \pmv{56.6}{0.6} & \pmv{-10.3}{1.4} \\

\textsc{KnOTS-TIES}
& \pmv{54.4}{6.9} & \pmv{22.5}{6.1} & \pmv{31.6}{7.2} & \pmv{-12.6}{3.9}
& \pmv{67.8}{0.4} & \pmv{\textbf{58.0}}{0.1} & \pmv{\underline{62.5}}{0.1} & \pmv{\textbf{1.9}}{0.5}
& \pmv{60.5}{1.4} & \pmv{51.5}{1.7} & \pmv{55.8}{1.5} & \pmv{\textbf{-1.3}}{0.7} \\

\textsc{TSV-M}
& \pmv{68.2}{4.8} & \pmv{50.0}{4.2} & \pmv{57.7}{4.0} & \pmv{-24.0}{5.6}
& \pmv{63.3}{4.8} & \pmv{50.3}{4.1} & \pmv{56.4}{4.3} & \pmv{-26.7}{4.9}
& \pmv{58.8}{3.3} & \pmv{51.5}{3.8} & \pmv{54.9}{3.5} & \pmv{-31.7}{3.4} \\

\textsc{OPCM}
& \pmv{75.5}{0.5} & \pmv{51.8}{0.4} & \pmv{61.4}{0.4} & \pmv{-6.3}{1.1}
& \pmv{71.9}{0.3} & \pmv{54.0}{0.1} & \pmv{61.7}{0.1} & \pmv{-6.0}{1.0}
& \pmv{65.7}{0.2} & \pmv{53.1}{0.3} & \pmv{\underline{58.6}}{0.3} & \pmv{-7.8}{1.5} \\

\textsc{NUFILT}
& \pmv{\textbf{83.6}}{0.2} & \pmv{52.3}{0.3} & \pmv{\underline{64.3}}{0.2} & \pmv{\textbf{-2.7}}{0.7}
& \pmv{\underline{78.0}}{0.2} & \pmv{49.4}{0.2} & \pmv{60.5}{0.2} & \pmv{-5.7}{0.9}
& \pmv{\underline{71.0}}{0.9} & \pmv{44.5}{0.4} & \pmv{54.8}{0.3} & \pmv{-8.9}{2.3} \\

\textbf{\methodshort{}}
& \pmv{\underline{83.0}}{0.3} & \pmv{\textbf{56.7}}{0.2} & \pmv{\textbf{67.4}}{0.2} & \pmv{\underline{-3.5}}{0.6}
& \pmv{\textbf{79.2}}{0.4} & \pmv{56.4}{0.2} & \pmv{\textbf{65.9}}{0.2} & \pmv{-4.8}{0.6}
& \pmv{\textbf{73.5}}{0.5} & \pmv{\textbf{54.9}}{0.4} & \pmv{\textbf{62.9}}{0.4} & \pmv{-8.2}{1.3} \\

\bottomrule
\end{tabular}%
}
\label{tab:b32_results}
\end{table*}

\subsection{Experimental Setups}
\textbf{Benchmarks and Protocols.}
Following \citep{ilharco2023editing, wanglocalizing}, we adopt CLIP-ViT backbones \citep{radford2021learning} and construct three groups of 8, 14, and 20 image classification tasks.
We use publicly available ViT-B/32, ViT-B/16, and ViT-L/14 checkpoints, each fine-tuned on up to 20 datasets \citep{tangFusionBenchComprehensiveBenchmark2024}.
To ensure robustness to task order, all vision experiments are repeated over 10 random permutations (seeds 42--51).
For NLP, we fine-tune Flan-T5-base \citep{chung2024scaling} on five GLUE tasks \citep{wang2019glue} (CoLA, MRPC, QQP, SST2, and STSB) and merge them once in alphabetical order; MNLI, QNLI, and RTE are held out as probes of general linguistic knowledge.
For vision, we additionally evaluate the merged model on three held-out control datasets: ImageNet (IN) \citep{deng2009imagenet}, ImageNet-R (IN-R) \citep{hendrycks2021many}, and ObjectNet (ObjNet) \citep{barbu2019objectnet}. These datasets assess whether the merged model retains the general visual representations of the pre-trained backbone.

\textbf{Implementation Details}
We apply risk-aware sparse masking and low-rank recovery to selected attention and first feed-forward linear layers.
Across all experiments, we share the subspace rank $r_p=128$, recovery rank $r_l=64$, and new-task recovery rank $r_v=8$ without task-specific tuning.
For ViT-B/32, the keep ratio is set to $\rho=0.5$ for 8-task merging and $\rho=0.3$ for 14- and 20-task merging. For ViT-B/16 and ViT-L/14, the corresponding values are $0.7$ and $0.5$. We use $\rho=0.9$ for Flan-T5 merging.
We set the plasticity weight to $\lambda=0.8$ and regularization weight to $\mu=0.1$, and optimize the recovery factors with Adam for 50 iterations per task using a learning rate of $1\times10^{-3}$.

\textbf{Metrics and Baselines.}
We evaluate performance with four metrics: average accuracy (ACC), backward transfer (BWT) \citep{lin2022beyond}, general performance average (Gen. Avg.), and their harmonic mean (H-score).
ACC is the mean accuracy of the final merged model across all tasks,
$
\text{ACC} = \tfrac{1}{T} \sum_{i=1}^{T} a_i(\theta_T^{\text{merged}}),
$
where \(a_i(\cdot)\) denotes accuracy on task \(i\).
BWT quantifies the effect of merging on past tasks by comparing their performance before and after the final merge:
$
\text{BWT} = \tfrac{1}{T-1} \sum_{i=1}^{T-1} \Big[a_i(\theta_T^{\text{merged}}) - a_i(\theta_i^{\text{merged}})\Big].
$
Gen. Avg. measures the average performance on the held-out control datasets, which probe whether the merged model retains the general knowledge of the pre-trained backbone:
$
\text{Gen. Avg.} = \tfrac{1}{|\mathcal{G}|}\textstyle\sum_{g \in \mathcal{G}} a_g,
$
where $\mathcal{G}$ is the set of held-out control datasets for each domain (IN, IN-R, and ObjNet for vision; MNLI, QNLI, and RTE for NLP).
H-score is the harmonic mean of ACC and Gen. Avg., emphasizing that a strong continual merging method must achieve both high merging accuracy and strong general performance:
$
\text{H-score} = \tfrac{2 \cdot \text{ACC} \cdot \text{Gen. Avg.}}{\text{ACC} + \text{Gen. Avg.}}.
$
We compare against Task Arithmetic \citep{ilharco2023editing},
\textsc{Ties-Merging} \citep{yadav2023ties}, \textsc{WUDI-Merging}
\citep{chengwhoever}, \textsc{Iso-C} \citep{marczak2025iso},
\textsc{KnOTS-TIES} \citep{stoica2025model}, \textsc{TSV-M}
\citep{gargiulo2025tsv}, \textsc{OPCM} \citep{tang2025merging}, and
\textsc{NUFILT} \citep{qiu2026nullspace}.
Detailed descriptions are provided in the supplementary material.

\subsection{Main Results}

\begin{table}[t]
\centering
\caption{Continual merging results on the 5-task NLP benchmark
(five GLUE tasks merged in alphabetical order). MNLI, QNLI, and RTE serve as
held-out probes of general performance (Gen.). ACC and BWT are computed on the
merged tasks, and H-score is the harmonic mean of ACC and Gen. Best results
are in \textbf{bold} and second-best are \underline{underlined}.}
\label{tab:nlp_results}
\renewcommand\arraystretch{1.12}
\setlength{\tabcolsep}{2pt}
\resizebox{\linewidth}{!}{%
\begin{tabular}{lccccccc}
\toprule
\textbf{Method} & \textbf{ACC$\uparrow$} & \textbf{MNLI} & \textbf{QNLI} & \textbf{RTE} & \textbf{Gen.$\uparrow$} & \textbf{H-score$\uparrow$} & \textbf{BWT$\uparrow$} \\
\midrule
Pre-Trained    & 76.2 & 56.5 & 88.4 & 80.1 & 75.0 & 75.6 & -- \\
Individual     & 86.0 & 47.8 & 87.7 & 77.8 & 71.1 & 77.9 & -- \\
\midrule
Task Arithmetic & 79.0 & \textbf{58.1} & 88.9 & 79.1 & \textbf{75.4} & 77.2 & \underline{$-$1.1} \\
Ties-Merging    & 79.7 & 54.2 & \textbf{89.5} & 79.4 & 74.4 & 77.0 & $-$1.5 \\
OPCM           & 80.6 & 54.8 & \underline{89.3} & \underline{80.5} & 74.8 & 77.6 & $-$1.2 \\
WUDI-Merging   & \underline{84.2} & 52.1 & 88.6 & 78.3 & 73.0 & 78.2 & $-$1.5 \\
NUFILT         & \textbf{84.4} & 52.9 & 88.5 & 80.1 & 73.9 & \underline{78.8} & \textbf{$-$0.6} \\
\textbf{\methodshort{}} & 84.1 & \underline{55.8} & 89.2 & \textbf{80.9} & \underline{75.3} & \textbf{79.4} & $-$1.4 \\
\bottomrule
\end{tabular}%
}
\end{table}

\textbf{Results on Vision Tasks}.
Tab.~\ref{tab:b32_results} summarizes the vision results. \methodshort{} achieves the best H-score at every sequence length, exceeding the strongest H-score baseline by 3.1, 3.4, and 4.3 points for 8, 14, and 20 tasks. More importantly, on 14 and 20 tasks, it improves over NUFILT, the strongest ACC baseline, by 1.2/2.5 ACC points and 5.4/8.1 H-score points. These simultaneous gains show that the advantage is balanced rather than driven by a single metric, and grows as interference accumulates.

\textbf{Results on NLP Tasks}.
Tab.~\ref{tab:nlp_results} reports the Flan-T5-base results. NUFILT attains the highest ACC and least-negative BWT, while Task Arithmetic best preserves general performance, but neither optimizes their balance. \methodshort{} instead achieves the best H-score (79.4), improving on NUFILT by 0.6 points while remaining within 0.3 points of its ACC and improving Gen. by 1.4 points; it also obtains the best RTE accuracy. Moreover, after absorbing five tasks, \methodshort{} keeps general performance slightly above the Pre-Trained reference, whereas individual fine-tuning erodes it by about four points. These results show that the H-score gain reflects balanced task acquisition and general-knowledge preservation.

\begin{figure}[t]
  \centering
  \captionsetup[subfigure]{labelfont=bf,justification=centering,singlelinecheck=true}
  \begin{subfigure}[t]{0.48\columnwidth}
    \centering
    \includegraphics[width=1\linewidth]{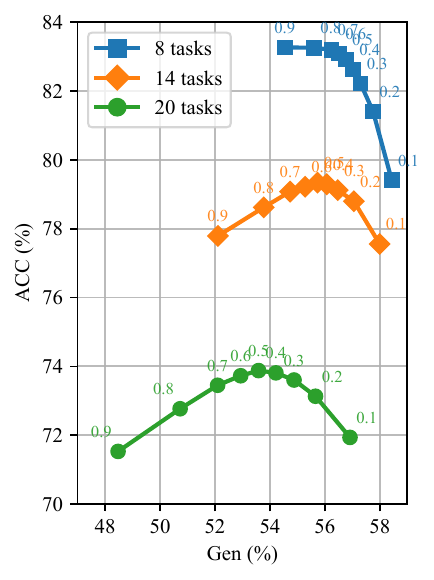}
    \caption{Mask keep ratio $\rho$}
    \label{fig:sensitivity_a}
  \end{subfigure}
  \begin{subfigure}[t]{0.48\columnwidth}
    \centering
    \includegraphics[width=1\linewidth]{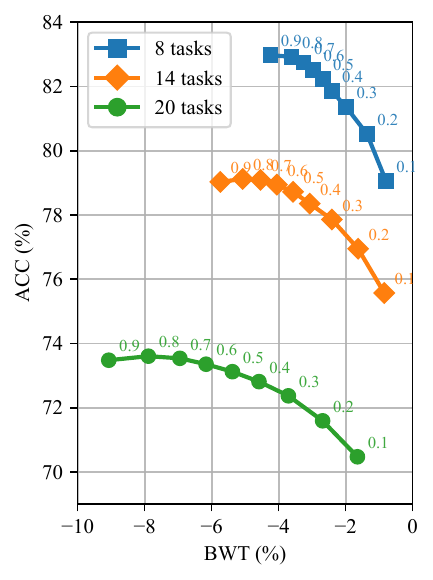}
    \caption{Plasticity weight $\lambda$}
    \label{fig:sensitivity_b}
  \end{subfigure}
  \caption{Analysis of risk-aware masking and stability--plasticity trade-off on CLIP ViT-B/32. (a) ACC--Gen trade-off when varying the mask keep ratio $\rho$ over 8, 14, and 20 tasks. (b) ACC--BWT trade-off when varying the plasticity weight $\lambda$ over 8, 14, and 20 tasks.}
  \label{fig:sensitivity}
\end{figure}

\subsection{Ablation and Analysis Results}

\begin{figure*}[t]
  \centering
  \captionsetup[subfigure]{labelfont=bf,justification=centering,singlelinecheck=true}
    \begin{subfigure}[t]{0.31\textwidth}
    \centering 
    \includegraphics[width=1\linewidth]{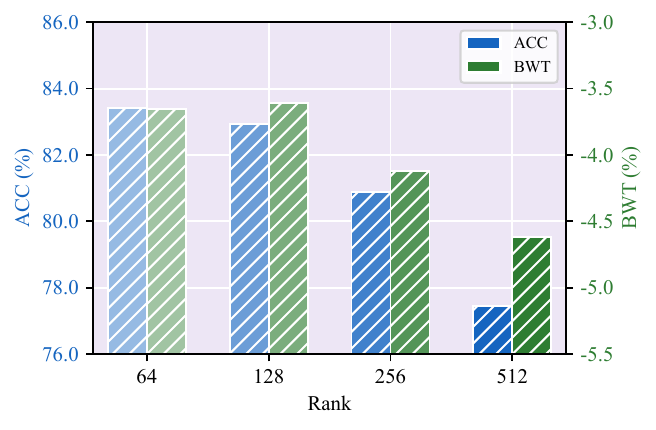}
    \caption{Scoring rank $r_p$}
  \end{subfigure}
  \hfill
  \begin{subfigure}[t]{0.31\textwidth}
    \centering
    \includegraphics[width=1\linewidth]{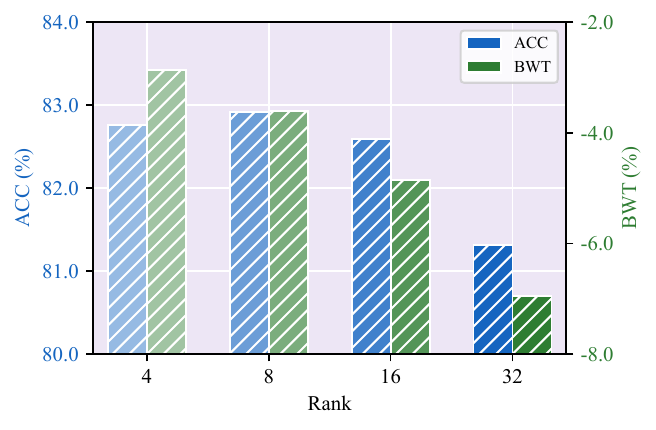}
    \caption{Task rank $r_v$}
  \end{subfigure}
  \hfill
  \begin{subfigure}[t]{0.31\textwidth}
    \centering
    \includegraphics[width=1\linewidth]{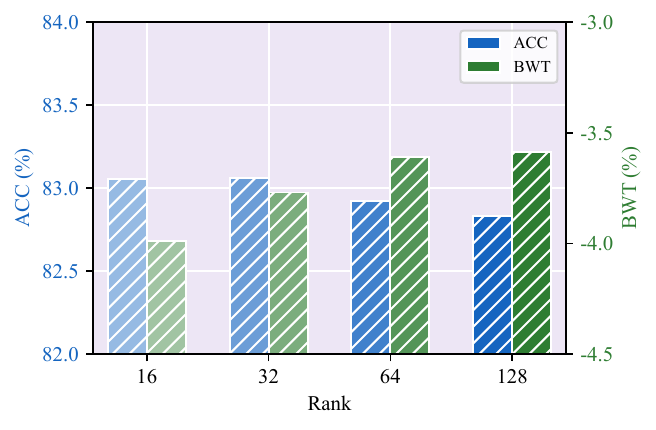}
    \caption{Recovery rank $r_l$}
  \end{subfigure}
  \caption{Rank sensitivity on CLIP ViT-B/32 under the 8-task continual merging protocol. We report ACC and BWT while varying the scoring rank $r_p$, task-subspace rank $r_v$, and masked low-rank recovery rank $r_l$.}
  \label{fig:tf_evaluation}
\end{figure*}

\begin{table*}[t]
\centering
\caption{Ablation study of \methodshort{} with CLIP ViT-B/32 over 8, 14, and 20 tasks. $M/R$ indicates whether risk-aware masking ($M_t$) and low-rank recovery ($B_tA_t$) are enabled.}
\label{tab:ablation}
\renewcommand\arraystretch{1.12}
\setlength{\tabcolsep}{4pt}
\newcommand{\abpm}[2]{\makecell{#1\\[-0.2ex]{\scriptsize$\pm$#2}}}
\resizebox{\textwidth}{!}{%
\begin{tabular}{l|c|ccc|ccc|ccc}
\toprule
\multirow{2}{*}{\textbf{Ablation}}
& \multirow{2}{*}{\textbf{Merged update $\Delta_t$}}
& \multicolumn{3}{c|}{\textbf{8 Tasks}}
& \multicolumn{3}{c|}{\textbf{14 Tasks}}
& \multicolumn{3}{c}{\textbf{20 Tasks}} \\
\cmidrule[0.5pt](lr){3-5} \cmidrule[0.5pt](lr){6-8} \cmidrule[0.5pt](lr){9-11}
& & \textbf{ACC$\uparrow$} & \textbf{Gen.$\uparrow$} & \textbf{BWT$\uparrow$} & \textbf{ACC$\uparrow$} & \textbf{Gen.$\uparrow$} & \textbf{BWT$\uparrow$} & \textbf{ACC$\uparrow$} & \textbf{Gen.$\uparrow$} & \textbf{BWT$\uparrow$} \\
\midrule
\textbf{Naive accumulation} \quad ($M/R$: \xmark/\xmark)
& $\tau_t$
& \abpm{62.0}{0.0} & \abpm{23.5}{0.0} & \abpm{-18.9}{5.3}
& \abpm{46.4}{0.0} & \abpm{19.1}{0.0} & \abpm{-25.8}{2.5}
& \abpm{34.2}{0.0} & \abpm{10.1}{0.0} & \abpm{-24.9}{5.4} \\
\textbf{Risk-aware mask only} \quad ($M/R$: \cmark/\xmark)
& $M_t\odot\tau_t$
& \abpm{80.7}{0.0} & \abpm{52.3}{0.0} & \abpm{-4.0}{0.9}
& \abpm{75.2}{0.0} & \abpm{52.7}{0.0} & \abpm{-5.6}{0.5}
& \abpm{63.3}{0.0} & \abpm{44.0}{0.0} & \abpm{-11.1}{2.2} \\
\textbf{Low-rank recovery only} \quad ($M/R$: \xmark/\cmark)
& $\tau_t+B_tA_t$
& \abpm{79.1}{1.8} & \abpm{47.2}{1.7} & \abpm{-10.1}{2.4}
& \abpm{54.2}{8.4} & \abpm{32.4}{5.5} & \abpm{-22.2}{10.0}
& \abpm{38.0}{13.7} & \abpm{19.5}{8.5} & \abpm{-34.0}{12.4} \\
\textbf{\methodshort{}} \quad ($M/R$: \cmark/\cmark)
& $M_t\odot(\tau_t+B_tA_t)$
& \abpm{83.0}{0.3} & \abpm{56.7}{0.2} & \abpm{-3.5}{0.6}
& \abpm{79.2}{0.4} & \abpm{56.4}{0.2} & \abpm{-4.8}{0.6}
& \abpm{73.5}{0.5} & \abpm{54.9}{0.4} & \abpm{-8.2}{1.3} \\
\bottomrule
\end{tabular}}
\end{table*}

\textbf{Ablation on Each Component.}
Tab.~\ref{tab:ablation} disentangles risk-aware masking and masked low-rank recovery. Naive accumulation collapses as the sequence grows, with ACC/Gen. falling from 62.0/23.5 at 8 tasks to 34.2/10.1 at 20 tasks. Masking provides the largest contribution: at 20 tasks it improves ACC/Gen. by 29.1/33.9 points and BWT from $-24.9$ to $-11.1$. Recovery alone is competitive at 8 tasks but degrades to 38.0 ACC and $-34.0$ BWT at 20 tasks with high variance, showing that an unconstrained correction can amplify interference. Combining both components performs best throughout; relative to masking alone, recovery adds 10.2/10.9 ACC/Gen. points at 20 tasks and makes BWT less negative. It also reduces the 20-task ACC/BWT standard deviations from 13.7/12.4 to 0.5/1.3, confirming robustness to task order. Thus, masking defines a safe support while recovery restores discarded task information, with their complementarity becoming more important in longer sequences.

\textbf{Risk and Trade-off Analysis.}
Fig.~\ref{fig:sensitivity} exposes two complementary trade-offs. In (a), decreasing the keep ratio $\rho$ consistently improves Gen., whereas ACC follows an inverted-U: retaining too many updates admits interference, while excessive sparsity removes task information. For 20 tasks, moving from $\rho=0.9$ to 0.5 raises ACC from about 71.5 to 73.9 and Gen. from 48.5 to 53.5, but $\rho=0.1$ reduces ACC below 72. The knees at $\rho=0.3$--$0.5$ motivate stronger sparsification for longer sequences. In (b), increasing $\lambda$ emphasizes new-task recovery, raising ACC while making BWT more negative; at 20 tasks, $\lambda=0.8$ gains roughly three ACC points over 0.1, with BWT changing from about $-1.6$ to $-8.0$. The same ordered trend at 8 and 14 tasks shows that $\lambda$ provides predictable control as interference accumulates. These curves support our global choice $\lambda=0.8$, with masking limiting its overwrite risk.

\textbf{Rank Sensitivity.}
Fig.~\ref{fig:tf_evaluation} shows structured rather than uniform rank sensitivity. Compact scoring subspaces work best: increasing $r_p$ from 64--128 to 512 lowers ACC from about 83 to 77.5 and also worsens BWT. Likewise, $r_v=8$ gives the highest ACC and $r_v=4$ the least-negative BWT, whereas $r_v=32$ degrades both metrics by overemphasizing incoming-task directions. In contrast, recovery is robust to $r_l$: across 16--128, ACC varies by less than 0.3 points and BWT slightly improves with rank. This contrast shows that subspace selectivity matters more than simply increasing correction capacity. The choices $r_p=128$, $r_v=8$, and $r_l=64$ therefore use compact scoring/task subspaces and place recovery in its broad stable region.

\begin{table}[t]
 \centering
\renewcommand\arraystretch{1.05}
\setlength{\tabcolsep}{4pt}
\caption{Overhead of ViT-B/32 under 8-task continual merging: per-task iterations, solving time, total time, and peak GPU memory.}
\label{tab:trainning_cost} 
\resizebox{\columnwidth}{!}{%
\begin{tabular}{lccccc}
\toprule
\textbf{Model} & \textbf{Iters.} & \textbf{Solv.} & \textbf{Tot.} & \textbf{GPU Mem.} & \textbf{H-score} \\
\midrule
\textsc{Task Arithmetic}  & - & - & 8.3s & - & 53.6\scriptsize{ $\pm$ 0.0} \\
\textsc{OPCM} & - & - & 76.6s & 1.3GB & 61.4\scriptsize{ $\pm$ 0.4} \\
\textsc{WUDI-Merging} & 50 & 28.9s & 37.7s & 2.0GB & 54.9\scriptsize{ $\pm$ 7.5} \\
\textsc{NUFILT} & 50 & 18.4s & 138.9s & 1.8GB & 64.3\scriptsize{ $\pm$0.2} \\
\textbf{\methodshort{}} & 50 & 21.3s & 216.3s & 2.3GB & 67.4\scriptsize{ $\pm$0.2} \\
\bottomrule
\end{tabular}}
\end{table}

\textbf{Computation Overhead.}
Tab.~\ref{tab:trainning_cost} shows that \methodshort{} uses 216.3s and 2.3GB for eight tasks. Compared with NUFILT, the additional 77.4s and 0.5GB yield a 3.1-point H-score gain (67.4 vs. 64.3), while training-free methods are faster but less accurate. This overhead is incurred only during offline merging: the recovery factors are folded into the backbone, adding no inference parameters or latency, and the one-time cost is amortized over subsequent deployment.

\section{Conclusion}
In this work, we addressed data-free continual model merging, where
independently fine-tuned models must be integrated sequentially without task
data or access to earlier checkpoints. The key challenge is to acquire new
knowledge while preserving both prior tasks and general
capabilities. To this end, we introduced \methodshort{}, which scores each
parameter update by balancing its potential to overwrite general knowledge
against its task-specific utility. Guided by this score, it retains safe
updates through sparse masking and recovers discarded task information with a
constrained low-rank correction. Experiments across vision and language
benchmarks show that \methodshort{} consistently achieves the best H-score,
demonstrating a stronger balance between task acquisition and general-knowledge
preservation.

\bibliography{aaai2027}

\clearpage
\appendix

\renewcommand{\theequation}{S.\arabic{equation}}
\setcounter{equation}{0}
\renewcommand{\thefigure}{S.\arabic{figure}}
\setcounter{figure}{0}
\renewcommand{\thetable}{S.\arabic{table}}
\setcounter{table}{0}

\renewcommand{\thetheorem}{A.\arabic{theorem}}
\setcounter{theorem}{0}

\renewcommand{\thelemma}{A.\arabic{lemma}}
\setcounter{lemma}{0}

\renewcommand{\theproposition}{A.\arabic{proposition}}
\setcounter{proposition}{0}

\renewcommand{\thecorollary}{A.\arabic{corollary}}
\setcounter{corollary}{0}

\section*{Appendix}
The appendix is organized into sections, each providing supplementary explanations and supporting details.

\begin{quote}
\noindent 
\textbf{A \quad Additional Descriptions} \dotfill \ref{appdix:Additional Descriptions} \\
\hspace{1em} A.1 \quad From Loss Curvature to Response Sensitivity \dotfill \ref{appdix:curvature-response} \\
\hspace{1em} A.2 \quad Recovery-Objective Derivation \dotfill \ref{appdix:recovery-derivation} \\
\hspace{1em} A.3 \quad Dataset Suite and Evaluation Protocol \dotfill \ref{appdix:Details of Dataset and Task Settings} \\
\hspace{1em} A.4 \quad Source Checkpoints and Fine-Tuning Setup \dotfill \ref{appdix:Details of Downstream Models} \\
\hspace{1em} A.5 \quad Baseline Implementations \dotfill \ref{appdix:Details of Baselines} \\
\hspace{1em} A.6 \quad Practical Hyperparameter Selection \dotfill \ref{appdix:Hyper-parameter guide} \\

\textbf{B \quad Additional Results} \dotfill \ref{appdix:Additional Results} \\
\hspace{1em} B.1 \quad Detailed Overall Performance Results \dotfill \ref{appdix:Detailed Overall Performance Results} \\

\textbf{C \quad Discussions} \dotfill \ref{appdix:Discussions} \\
\hspace{1em} C.1 \quad Limitations \dotfill \ref{appdix:Limitations} \\
\hspace{1em} C.2 \quad Broader Impacts \dotfill \ref{appdix:Broader Impacts} \\
\hspace{1em} C.3 \quad LLM Usage
\end{quote}

\section{Additional Descriptions}
\label{appdix:Additional Descriptions}

\subsection{From Loss Curvature to Response Sensitivity}
\label{appdix:curvature-response}
We make explicit the connection between Eq.~(1) and Eq.~(2) of the main paper. For a perturbation $\Delta W^{(l)}$ to a linear layer with input $x$, let $C_g^{(l)}=\mathbb E[xx^\top]$. The squared response-change proxy can be written as
\begin{align*}
\mathcal R_g^{(l)}(\Delta W)
&=\tfrac12\mathbb E\|\Delta W^{(l)}x\|_2^2 \\
&=\tfrac12\operatorname{vec}(\Delta W^{(l)})^\top
\bigl(C_g^{(l)}\otimes I\bigr)
\operatorname{vec}(\Delta W^{(l)}).
\end{align*}
Its gradient vanishes at $\Delta W^{(l)}=0$, and its Hessian with respect to $\operatorname{vec}(\Delta W^{(l)})$ is $C_g^{(l)}\otimes I$. If only entry $(i,j)$ changes by $\delta_{ij}$, then $\mathcal R_g^{(l)}=\tfrac12[C_g^{(l)}]_{jj}\delta_{ij}^2$, matching the quadratic form in Eq.~(1). Thus, if response preservation proxies general-knowledge preservation, $C_g^{(l)}\otimes I$ is a tractable layer-wise surrogate---not an exact identity---for the unknown loss Hessian. Without data, the main paper estimates the dominant input-side subspace of $C_g^{(l)}$ from the pretrained weights.

\subsection{Data-Free Derivation of the Recovery Objectives}
\label{appdix:recovery-derivation}
We derive Eq.~(10) and Eq.~(11) of the main paper from response constraints that would be available if task data could be accessed. We omit the layer superscript for readability. For $s\in\{o,n\}$, construct $X_s=[x_{s,1},\ldots,x_{s,N_s}]^\top\in\mathbb R^{N_s\times d_i}$ by stacking layer-input representations from earlier tasks ($s=o$) or the incoming task ($s=n$). Thus, the columns of $DX_s^\top=[Dx_{s,1},\ldots,Dx_{s,N_s}]$ are the per-sample response changes induced by a weight difference $D$. Let $\bar\tau_{t-1}=\theta_{t-1}^{\mathrm{merged}}-\theta_0$ and let $\Delta_t$ be the recovered update. For a shared layer input, the weight differences governing the old- and new-task response errors are
\begin{align*}
D_{\mathrm{old}}
&=(\theta_{t-1}^{\mathrm{merged}}+\Delta_t)
-\theta_{t-1}^{\mathrm{merged}}=\Delta_t,\\
D_{\mathrm{new}}
&=(\theta_0+\bar\tau_{t-1}+\Delta_t)
-(\theta_0+\tau_t)
=\bar\tau_{t-1}+\Delta_t-\tau_t.
\end{align*}
Using squared response distance and omitting dataset-size constants, the corresponding data-dependent losses are
\begin{align}
\mathcal J_{\mathrm{old}}^{\mathrm{data}}
&=\|D_{\mathrm{old}}X_o^\top\|_F^2,
\label{eq:app-data-old}\\
\mathcal J_{\mathrm{new}}^{\mathrm{data}}
&=\|D_{\mathrm{new}}X_n^\top\|_F^2.
\label{eq:app-data-new}
\end{align}
The first loss keeps old-task layer responses unchanged, whereas the second matches the response of the individually fine-tuned incoming-task model. Neither can be evaluated in the data-free setting because $X_o$ and $X_n$ are unavailable.

\paragraph{Upper bound from prior work.}
Corollary 1 of \citet{qiu2026nullspace} directly gives the data-free inequality used here. For a rank-$r_d$ representation matrix $X$, the parameter-derived right singular basis $\widehat V$, and any $D\in\mathbb R^{d_o\times d_i}$, the result states
\begin{equation}
\|DX^\top\|_F^2
\leq 2\sigma_1(X)^2\left(
\|D\widehat V\|_F^2+r_d\zeta^2\|D\|_2^2
\right).
\label{eq:app-projection-bound}
\end{equation}
Here $\sigma_1(X)$ is the largest singular value of $X$, while $\zeta$ is the representation--parameter subspace misalignment term defined in that work. We use this previously established result under its stated conditions and only specialize it to the SAFE-Merge recovery variables below.

We use $V_{\mathrm{old}}$ from $\bar\tau_{t-1}$ as the proxy for the dominant right subspace of $X_o$, and $V_{\mathrm{new}}$ from $\tau_t$ as the corresponding proxy for $X_n$. Applying Eq.~(\ref{eq:app-projection-bound}) to Eq.~(\ref{eq:app-data-old}) and Eq.~(\ref{eq:app-data-new}) yields
\begin{align}
\mathcal J_{\mathrm{old}}^{\mathrm{data}}
&\leq 2\sigma_1(X_o)^2\left(
\|D_{\mathrm{old}}V_{\mathrm{old}}\|_F^2
+r_o\zeta_o^2\|D_{\mathrm{old}}\|_2^2\right),\label{eq:app-old-bound}\\
\mathcal J_{\mathrm{new}}^{\mathrm{data}}
&\leq 2\sigma_1(X_n)^2\left(
\|D_{\mathrm{new}}V_{\mathrm{new}}\|_F^2
+r_n\zeta_n^2\|D_{\mathrm{new}}\|_2^2\right).
\label{eq:app-new-bound}
\end{align}
Because $G=VV^\top$ and $V^\top V=I$, $\|DG\|_F^2=\|DV\|_F^2$. The first terms on the right of Eq.~(\ref{eq:app-old-bound}) and Eq.~(\ref{eq:app-new-bound}) are therefore exactly the objectives in Eq.~(10) and Eq.~(11) of the main paper.

\paragraph{Connection to the recovery regularizer.}
The projection terms in Eq.~(\ref{eq:app-old-bound}) and Eq.~(\ref{eq:app-new-bound}) give Eq.~(10) and Eq.~(11), while each bound also contains a spectral-norm residual. In both residuals, the only trainable component is the masked recovery $M_t\odot(B_tA_t)$; all remaining terms are fixed during recovery. Using $\|A+B\|_2^2\leq2\|A\|_2^2+2\|B\|_2^2$ and $\|B\|_2\leq\|B\|_F$, each residual is bounded by a fixed term plus $2\|M_t\odot(B_tA_t)\|_F^2$. The latter is exactly the regularizer in Eq.~(12). Together, Eq.~(10), Eq.~(11), and Eq.~(12) minimize the optimizable part of the two upper bounds, with the unavailable data-dependent scales absorbed into the loss weights. The result motivates the recovery objective rather than providing a bound that can be evaluated without data.

\subsection{Dataset Suite and Evaluation Protocol}
\label{appdix:Details of Dataset and Task Settings}
\paragraph{Vision benchmark collection.}
Our vision evaluation uses 20 classification datasets drawn from markedly different domains, including natural images, satellite scenes, pathology patches, facial expressions, handwritten characters, and rendered text. Following the broad multi-domain protocol of prior continual-merging work~\citep{tang2025merging}, the collection also spans substantially different label granularities, from binary decisions to problems with hundreds of classes. The complete suite consists of
SUN397~\citep{xiao2010sun}, 
Stanford Cars~\citep{krause20133d}, 
RESISC45~\citep{cheng2017remote}, 
EuroSAT~\citep{helber2019eurosat}, 
SVHN~\citep{netzer2011reading}, 
GTSRB~\citep{stallkamp2012man}, 
MNIST~\citep{lecun1998mnist}, 
DTD~\citep{cimpoi2014describing}, 
Flowers102~\citep{nilsback2008automated}, 
PCAM~\citep{veeling2018rotation}, 
FER2013~\citep{goodfellow2013challenges}, 
Oxford-IIIT Pet~\citep{parkhi2012cats}, 
STL-10~\citep{coates2011analysis}, 
CIFAR-100 and CIFAR-10~\citep{krizhevsky2009learning}, 
Food-101~\citep{bossard2014food}, 
Fashion-MNIST~\citep{xiao2017fashion}, 
EMNIST~\citep{cohen2017emnist}, 
KMNIST~\citep{clanuwat2018deep}, 
and Rendered SST2~\citep{socher2013recursive}.

\paragraph{Language benchmark collection.}
For language, we adopt eight datasets from GLUE~\citep{wang2019glue}, covering grammatical acceptability, inference, paraphrase recognition, question-pair matching, sentiment, and semantic similarity. Exact-match accuracy is used for the classification tasks: CoLA, MNLI, MRPC, QNLI, QQP, RTE, and SST2. We evaluate STSB, the semantic-similarity task, with Spearman's $\rho$.

\paragraph{Nested vision streams.}
The vision datasets form three nested streams containing 8, 14, or 20 tasks. We measure each stream with average accuracy (ACC) and backward transfer (BWT). For every stream length, ten independently sampled task permutations are evaluated, and all reported results aggregate their mean and standard deviation; the exact permutations appear in Tab.~\ref{tab:appendix_dataset_order}.

\begin{itemize}
    \item \textbf{8-task stream}:  
    (1) SUN397, (2) Stanford Cars, (3) RESISC45, (4) EuroSAT, (5) SVHN, (6) GTSRB, (7) MNIST, (8) DTD.
    
    \item \textbf{14-task stream}:  
    The preceding eight datasets are followed by
    (9) Flowers102, (10) PCAM, (11) FER2013, (12) Oxford-IIIT Pet, (13) STL-10, (14) CIFAR-100.
    
    \item \textbf{20-task stream}:  
    The remaining six datasets extend the 14-task stream:
    (15) CIFAR-10, (16) Food-101, (17) Fashion-MNIST, (18) EMNIST, (19) KMNIST, (20) Rendered SST2.
\end{itemize}

The language protocol uses one deterministic sequence: CoLA, MRPC, QQP, SST2, and STSB are merged in alphabetical order. MNLI, QNLI, and RTE are excluded from the merge stream and used exclusively to probe retained general linguistic knowledge.

\newcommand{\arr}{$\,\rightarrow\,$}   
\begin{table*}[h]
\centering
\setlength{\tabcolsep}{3pt}
\renewcommand{\arraystretch}{1.3}
\caption{Task permutations evaluated for the 8-, 14-, and 20-dataset vision streams.}
\label{tab:task_orderings}
\begin{tabular}{p{0.4cm}|c|l}
\toprule
 & \textbf{Order} & \textbf{Dataset Order (by ID)} \\
\midrule
\multirow{10}{*}{\rotatebox[origin=c]{90}{\textbf{\small{8 tasks}}}} 
& 1  & (04\arr05\arr07\arr08\arr03\arr06\arr01\arr02)\\
& 2  & (07\arr08\arr05\arr04\arr02\arr06\arr03\arr01)\\
& 3  & (03\arr06\arr04\arr02\arr01\arr08\arr05\arr07)\\
& 4  & (06\arr08\arr02\arr01\arr03\arr07\arr04\arr05)\\
& 5  & (07\arr06\arr03\arr08\arr05\arr01\arr04\arr02)\\
& 6  & (07\arr02\arr03\arr08\arr05\arr04\arr01\arr06)\\
& 7  & (07\arr01\arr04\arr03\arr08\arr05\arr02\arr06)\\
& 8  & (08\arr05\arr06\arr07\arr01\arr04\arr03\arr02)\\
& 9  & (01\arr04\arr05\arr02\arr06\arr03\arr07\arr08)\\
& 10 & (08\arr03\arr01\arr02\arr06\arr05\arr07\arr04)\\
\midrule
\multirow{10}{*}{\rotatebox[origin=c]{90}{\textbf{\small{14 tasks}}}} 
& 1  & (09\arr13\arr08\arr07\arr14\arr12\arr06\arr03\arr10\arr04\arr05\arr01\arr02\arr11)\\
& 2  & (09\arr10\arr11\arr14\arr07\arr13\arr04\arr02\arr06\arr08\arr03\arr12\arr05\arr01)\\
& 3  & (05\arr08\arr12\arr06\arr11\arr01\arr10\arr04\arr14\arr03\arr02\arr13\arr09\arr07)\\
& 4  & (03\arr10\arr09\arr12\arr04\arr13\arr01\arr06\arr11\arr02\arr14\arr08\arr07\arr05)\\
& 5  & (08\arr14\arr09\arr06\arr12\arr13\arr05\arr03\arr04\arr11\arr10\arr01\arr07\arr02)\\
& 6  & (03\arr12\arr13\arr01\arr11\arr04\arr10\arr05\arr14\arr08\arr09\arr07\arr02\arr06)\\
& 7  & (07\arr01\arr12\arr10\arr02\arr08\arr13\arr04\arr05\arr11\arr14\arr03\arr06\arr09)\\
& 8  & (05\arr12\arr04\arr11\arr03\arr08\arr10\arr01\arr09\arr13\arr14\arr07\arr06\arr02)\\
& 9  & (10\arr07\arr09\arr02\arr03\arr13\arr01\arr12\arr14\arr04\arr11\arr06\arr05\arr08)\\
& 10 & (01\arr02\arr11\arr06\arr08\arr12\arr07\arr05\arr10\arr14\arr03\arr13\arr09\arr04)\\
\midrule
\multirow{10}{*}{\rotatebox[origin=c]{90}{\textbf{\small{20 tasks}}}} 
& 1  & (20\arr06\arr15\arr05\arr10\arr14\arr16\arr19\arr07\arr13\arr18\arr11\arr02\arr12\arr03\arr17\arr08\arr09\arr01\arr04)\\
& 2  & (09\arr14\arr06\arr03\arr07\arr04\arr18\arr01\arr17\arr19\arr08\arr20\arr13\arr16\arr11\arr12\arr15\arr05\arr10\arr02)\\
& 3  & (09\arr15\arr16\arr11\arr03\arr13\arr08\arr10\arr12\arr02\arr20\arr01\arr05\arr19\arr07\arr06\arr04\arr18\arr17\arr14)\\
& 4  & (17\arr04\arr11\arr19\arr18\arr10\arr07\arr15\arr12\arr13\arr08\arr02\arr01\arr06\arr05\arr03\arr20\arr16\arr14\arr09)\\
& 5  & (14\arr16\arr04\arr20\arr15\arr17\arr07\arr11\arr06\arr18\arr12\arr01\arr19\arr09\arr10\arr05\arr08\arr02\arr13\arr03)\\
& 6  & (02\arr06\arr17\arr04\arr19\arr18\arr08\arr16\arr20\arr01\arr10\arr13\arr07\arr09\arr05\arr11\arr15\arr14\arr03\arr12)\\
& 7  & (19\arr01\arr09\arr14\arr06\arr20\arr17\arr04\arr08\arr02\arr15\arr03\arr16\arr13\arr12\arr07\arr10\arr05\arr11\arr18)\\
& 8  & (15\arr07\arr08\arr02\arr10\arr06\arr17\arr20\arr05\arr19\arr16\arr01\arr18\arr09\arr13\arr11\arr04\arr14\arr12\arr03)\\
& 9  & (10\arr05\arr07\arr11\arr01\arr03\arr17\arr15\arr18\arr04\arr14\arr19\arr02\arr06\arr13\arr20\arr08\arr12\arr09\arr16)\\
& 10 & (01\arr11\arr02\arr15\arr03\arr10\arr12\arr19\arr16\arr13\arr07\arr05\arr09\arr04\arr14\arr20\arr06\arr18\arr17\arr08)\\
\bottomrule
\end{tabular}
\label{tab:appendix_dataset_order}
\end{table*}

\subsection{Source Checkpoints and Fine-Tuning Setup}
\label{appdix:Details of Downstream Models} 
We next specify the pretrained references, task-specialized checkpoints, and optimization recipes used to construct the merge streams in both modalities.

For vision, Tab.~\ref{tab:single_model_vit} lists the zero-shot CLIP ViT references together with the test accuracy of every dataset-specific checkpoint. We obtain the specialized models from Hugging Face (\url{https://huggingface.co/tanganke}). Their training updates only the image encoder and leaves the text encoder frozen. All checkpoints use cross-entropy, Adam, cosine learning-rate decay, a learning rate of $1\times10^{-5}$, batch size 128, and 4000 optimization steps.

The language experiments start from Flan-T5-base. Five GLUE checkpoints enter the continual merge, whereas MNLI, QNLI, and RTE remain held-out controls. Tab.~\ref{tab:single_model_flan_t5} records the pretrained reference and individual fine-tuning performance on all eight datasets. Each task model is optimized for 2000 steps with learning rate $4\times10^{-5}$ and batch size 16.

\begin{table*}[t]
    \centering
    \fontsize{8}{12}\selectfont  
    \setlength{\tabcolsep}{1pt}  
    \caption{Zero-shot CLIP accuracy and individual fine-tuning accuracy for the 20 vision datasets.}
    \label{tab:single_model_vit}
    \begin{tabular}{p{0.5cm}lcccccccccc}  
    \toprule
        & \textbf{Model}  & \rot{\scriptsize{SUN397}} & \rot{\scriptsize{Cars}} & \rot{\scriptsize{RESISC45}} & \rot{\scriptsize{EuroSAT}} & \rot{\scriptsize{SVHN}} & \rot{\scriptsize{GTSRB}} & \rot{\scriptsize{MNIST}} & \rot{\scriptsize{DTD}} & \rot{\scriptsize{Flowers102}} & \rot{\scriptsize{PCAM}} \\ \midrule
        \multirow{3}{*}{\rotatebox[origin=c]{90}{\textbf{\scriptsize{Pre-Trained}}}} & 
        CLIP ViT-B/32        & 63.2 & 59.6 & 60.3 & 45.0 & 31.6 & 32.5 & 48.3 & 44.2 & 66.4 & 60.6 \\
        &CLIP ViT-B/16        & 65.5 & 64.7 & 66.4 & 54.1 & 52.0 & 43.5 & 51.7 & 45.0 & 71.3 & 54.0 \\
        &CLIP ViT-L/14        & 68.2 & 77.9 & 71.3 & 61.2 & 58.4 & 50.5 & 76.3 & 55.5 & 79.2 & 51.2 \\
    \hline
        \multirow{3}{*}{\rotatebox[origin=c]{90}{\textbf{\scriptsize{Fine-tuned}}}} 
        & CLIP ViT-B/32        & 74.9 & 78.5 & 95.1 & 99.1 & 97.3 & 98.9 & 99.6 & 79.7 & 88.6 & 88.0 \\
        & CLIP ViT-B/16        & 78.9 & 85.9 & 96.6 & 99.0 & 97.6 & 99.0 & 99.7 & 82.3 & 94.9 & 90.6 \\
        & CLIP ViT-L/14        & 82.8 & 92.8 & 97.4 & 99.1 & 97.9 & 99.2 & 99.8 & 85.5 & 97.7 & 91.1 \\
    \midrule
        & \textbf{Model}  & \rot{\scriptsize{FER2013}} & \rot{\scriptsize{OxfordIIITPet}} & \rot{\scriptsize{STL10}} & \rot{\scriptsize{CIFAR100}} & \rot{\scriptsize{CIFAR10}} & \rot{\scriptsize{Food101}} & \rot{\scriptsize{FashionMNIST}} & \rot{\scriptsize{EMNIST}} & \rot{\scriptsize{KMNIST}} & \rot{\scriptsize{RenderedSST2}} \\ \midrule
        \multirow{3}{*}{\rotatebox[origin=c]{90}{\textbf{\scriptsize{Pre-Trained}}}} & 
        CLIP ViT-B/32        & 41.3 & 83.3 & 97.1 & 63.7 & 89.8 & 82.4 & 63.0 & 12.0 & 10.0 & 58.6 \\
        &CLIP ViT-B/16        & 46.4 & 88.4 & 98.3 & 66.3 & 90.8 & 87.0 & 67.3 & 12.4 & 11.2 & 60.6 \\
        &CLIP ViT-L/14        & 50.0 & 93.2 & 99.4 & 75.1 & 95.6 & 91.2 & 67.0 & 12.3 & 9.7 & 68.9 \\
    \midrule
        \multirow{3}{*}{\rotatebox[origin=c]{90}{\textbf{\scriptsize{Fine-tuned}}}} &
        CLIP ViT-B/32        & 71.6 & 92.5 & 97.5 & 88.4 & 97.6 & 88.4 & 94.7 & 95.6 & 98.2 & 71.3 \\
        &CLIP ViT-B/16        & 72.8 & 94.5 & 98.2 & 88.8 & 98.3 & 91.9 & 94.5 & 95.3 & 98.1 & 75.7 \\
        &CLIP ViT-L/14        & 75.9 & 95.7 & 99.2 & 93.0 & 99.1 & 94.8 & 95.3 & 95.4 & 98.3 & 80.5 \\
    \bottomrule
    \end{tabular}
\end{table*}

\begin{table*}[tb]
    \centering
    \fontsize{9}{16}\selectfont
    \setlength{\tabcolsep}{6pt}  
    \caption{Flan-T5-base performance before adaptation and after separate fine-tuning on each GLUE dataset.}
    \label{tab:single_model_flan_t5} 
    \begin{tabular}{lcccccccc}  
    \toprule
        \textbf{Model}     & \rot{\scriptsize{CoLA}} & \rot{\scriptsize{MNLI}} & \rot{\scriptsize{MRPC}} & \rot{\scriptsize{QNLI}} & \rot{\scriptsize{QQP}} & \rot{\scriptsize{RTE}} & \rot{\scriptsize{SST2}} & \rot{\scriptsize{STSB}} \\ \midrule
        Flan-T5-base (Pre-Trained)         &69.1&56.5&76.2&88.4&82.1&80.1&91.2&62.2 \\
    \midrule
        Flan-T5-base (Fine-tuned)         & 75.0&83.4&87.5&91.5&85.4&85.9&93.6&88.7\\
    \bottomrule
    \end{tabular}
\end{table*}

\subsection{Baseline Implementations}
\label{appdix:Details of Baselines}
The compared methods are instantiated as summarized below.
\begin{itemize}
    \item \textbf{Task Arithmetic.}
    Each specialized checkpoint is converted into a displacement from the common pretrained model. Continual composition then adds a scaled version of the incoming displacement to the current merged parameters~\citep{ilharco2023editing}.

    \item \textbf{Ties-Merging.}  
    Magnitude trimming and sign reconciliation are added to task-vector composition to limit redundant or contradictory coordinates~\citep{yadav2023ties}. At step $t$, the incoming vector $\tau_t = \theta_t - \theta_0$ is combined with the accumulated Ties vector as
    $\tau^{\text{Ties}}_t = \mathrm{Ties}\!\left(\tau^{\text{Ties}}_{t-1},\, \tau_t\right)$.
    The resulting parameters are
    $\theta^{\text{merged}}_t = \theta^{\text{merged}}_{t-1} + \lambda \tau^{\text{Ties}}_t$.
    
    \item \textbf{WUDI-Merging.}  
    This data-free baseline constrains residual merge errors to be orthogonal to the task directions~\citep{chengwhoever}. For linear layer $l$, we form the history vector $\tau^{\text{merged},(l)}_{t-1}=\theta^{\text{merged},(l)}_{t-1}-\theta_0^{(l)}$ and the new vector $\tau_t^{(l)}=\theta_t^{(l)}-\theta_0^{(l)}$. Its layerwise merged vector is optimized with
    \(\mathcal{L} = \sum_i \frac{1}{\|\tau_{i,l}\|^2_F} \big\| (\tau_{m,l} - \tau_{i,l})(\tau_{i,l})^\top \big\|_F^2\).
    After gradient-based optimization, the merged checkpoint is reconstructed as $\theta^{\text{merged}}_t=\theta_0+\tau_m$.

    \item \textbf{Iso-C.}
    Shared structure is separated from task-dependent subspaces, after which isotropic merging prevents individual tasks from dominating the composition~\citep{marczak2025iso}.

    \item \textbf{KnOTS-TIES.}
    Task updates are first mapped into an SVD-derived aligned basis, followed by Ties-Merging over the aligned factors~\citep{stoica2025model}.

    \item \textbf{TSV-M.}
    Model displacements are represented by compact task singular vectors, whose structured composition is designed to mitigate interference~\citep{gargiulo2025tsv}.

    \item \textbf{OPCM.}
    Each arriving update is projected relative to directions stored by the running model, suppressing components that would conflict with accumulated tasks~\citep{tang2025merging}.

    \item \textbf{NUFILT.}
    Earlier knowledge is protected through null-space filtering, while a projection-aware adaptation stage restores plasticity for the incoming task~\citep{qiu2026nullspace}.

\end{itemize}

\paragraph{Configuration used in comparison.}
\label{appdix:Details of Baseline Hyper-parameters}
For \methodshort{}, we use a learning rate of $10^{-3}$, 50 optimization steps, $(r_p,r_l,r_v)=(128,64,8)$, and $(\lambda,\mu)=(0.8,0.1)$ in all CLIP experiments. We set $\rho$ to $0.5$ and $0.3$ for ViT-B/32 on the 8-task and 14/20-task streams, respectively, and to $0.7$ and $0.5$ for ViT-B/16 and ViT-L/14 under the same two stream settings. These configurations are summarized in Tab.~\ref{tab:hyper-parameters}.

\begin{table*}[htbp]
  \centering
  \caption{Configurations used for CLIP continual merging; ``--'' marks parameters not used by a method.}
  \label{tab:hyper-parameters} 
    \fontsize{8}{14}\selectfont  
    \setlength{\tabcolsep}{4pt}  
  \resizebox{\textwidth}{!}{%
  \begin{tabular}{lllccccccccccc}
    \toprule
    Method & Backbone & Tasks & Scale Factor & $\alpha$ & Top-k (\%) & LR & Steps & $r_p$ & $r_l$ & $r_v$ & $\rho$ & $\lambda$ & $\mu$ \\
    \midrule
    \multirow{2}{*}{Task Arithmetic} & \multirow{2}{*}{All} & 8 & 0.3 & -- & -- & -- & -- & -- & -- & -- & -- & -- & -- \\
                                       & & 14/20 & 0.1 & -- & -- & -- & -- & -- & -- & -- & -- & -- & -- \\
     
    \multirow{2}{*}{Ties-Merging} & \multirow{2}{*}{All} & 8 & 0.3 & -- & 20 & -- & -- & -- & -- & -- & -- & -- & -- \\
                                       & & 14/20 & 0.1 & -- & 20 & -- & -- & -- & -- & -- & -- & -- & -- \\

    WUDI-Merging & All & 8/14/20 & -- & -- & -- & 1e-5 & 50 & -- & -- & -- & -- & -- & -- \\
    Iso-C & All & 8/14/20 & -- & -- & -- & -- & -- & -- & -- & -- & -- & -- & -- \\
    \multirow{2}{*}{KnOTS-TIES} & \multirow{2}{*}{All} & 8 & 0.3 & -- & 20 & -- & -- & -- & -- & -- & -- & -- & -- \\
                                       & & 14/20 & 0.1 & -- & 20 & -- & -- & -- & -- & -- & -- & -- & -- \\
    TSV-M & All & 8/14/20 & -- & -- & -- & -- & -- & -- & -- & -- & -- & -- & -- \\
    OPCM & All & 8/14/20 & -- & 0.5 & -- & -- & -- & -- & -- & -- & -- & -- & -- \\
    \multirow{2}{*}{NUFILT} & \multirow{2}{*}{All} & 8 & -- & -- & -- & 1e-3 & 50 & 128 & 64 & 8 & -- & -- & -- \\
                                       & & 14/20 & -- & -- & -- & 1e-3 & 50 & 128 & 64 & 8 & -- & -- & -- \\
    \multirow{4}{*}{\textbf{\methodshort{}}} & \multirow{2}{*}{ViT-B/32} & 8 & -- & -- & -- & 1e-3 & 50 & 128 & 64 & 8 & 0.5 & 0.8 & 0.1 \\
                                       & & 14/20 & -- & -- & -- & 1e-3 & 50 & 128 & 64 & 8 & 0.3 & 0.8 & 0.1 \\
                                       & \multirow{2}{*}{ViT-B/16, L/14} & 8 & -- & -- & -- & 1e-3 & 50 & 128 & 64 & 8 & 0.7 & 0.8 & 0.1 \\
                                       & & 14/20 & -- & -- & -- & 1e-3 & 50 & 128 & 64 & 8 & 0.5 & 0.8 & 0.1 \\
    \bottomrule
  \end{tabular}
  }
\end{table*}

\subsection{Practical Hyperparameter Selection}
\label{appdix:Hyper-parameter guide}
We summarize the role of each \methodshort{} hyperparameter and the configurations used in our experiments.

\paragraph{Subspace and recovery ranks $(r_p,r_l,r_v)$.}
The three ranks control different parts of the method:
\begin{itemize}
    \item \textbf{$r_p$} sets the dimension of the parameter subspace used for risk scoring and the representation of previously merged knowledge.
    \item \textbf{$r_l$} determines the rank, and therefore the capacity, of the masked recovery update.
    \item \textbf{$r_v$} specifies how many incoming-task directions participate in the projection-based recovery objective.
\end{itemize}
The rank analysis in the main paper shows that $r_p$ and $r_v$ are more sensitive than $r_l$, whose performance remains stable over the evaluated range. We therefore use $(r_p,r_l,r_v)=(128,64,8)$ throughout all experiments.

\paragraph{Sparsity and loss weights.}
The keep ratio $\rho$ directly sets the fraction of incoming coordinates retained by the mask. We choose $0.5$ and $0.3$ for ViT-B/32 under the 8-task and 14/20-task protocols, respectively. The corresponding values for ViT-B/16 and ViT-L/14 are $0.7$ and $0.5$, while Flan-T5 uses $0.9$. In the recovery objective, $\lambda$ controls the emphasis on learning the new task relative to preservation, and $\mu$ penalizes the low-rank correction. We fix $(\lambda,\mu)=(0.8,0.1)$ throughout; the main-paper sensitivity study further examines the effects of $\rho$ and $\lambda$.

\paragraph{Optimization length.}
We optimize the recovery factors for 50 steps with a learning rate of $10^{-3}$ in every experiment, matching the configuration reported in Tab.~\ref{tab:hyper-parameters}.

\paragraph{Transfer across model sizes.}
The same ranks, loss weights, and optimization settings are used for ViT-B/32, ViT-B/16, and ViT-L/14. Only the keep ratio is adjusted by backbone and stream length, as specified above.

\section{Additional Results}
\label{appdix:Additional Results}
This section provides detailed per-task results that complement the aggregate
metrics reported in the main paper.

\subsection{Detailed Overall Performance Results}
\label{appdix:Detailed Overall Performance Results}
Tab.~\ref{tab:appendix_b16_l14_results} reports aggregate continual merging results for ViT-B/16 and ViT-L/14, complementing the ViT-B/32 results in the main paper. Across both backbones and all sequence lengths, \methodshort{} achieves the highest H-score, consistently balancing merged-task accuracy with held-out general performance.

\begin{table*}[!htbp]
\centering
\caption{
Results on ViT-B/16 and ViT-L/14 over ten task orders. We report continual merging accuracy
(ACC) and backward transfer (BWT) on the merged tasks, general performance
(Gen.) on held-out control datasets, and their harmonic mean (H-score).
Among merging methods, best results are in \textbf{bold} and second-best are \underline{underlined}.
}
\begingroup
\setlength{\tabcolsep}{3.2pt}
\renewcommand\arraystretch{1.18}
\newcommand{\pmv}[2]{#1{\scriptsize\textsubscript{\ensuremath{\pm#2}}}}
\resizebox{\textwidth}{!}{%
\begin{tabular}{l|cccc|cccc|cccc}
\toprule
\multicolumn{13}{c}{\textbf{ViT-B/16}} \\
\hline
\multicolumn{1}{l|}{\textbf{Method}}
& \multicolumn{4}{c|}{\textbf{8 Tasks}}
& \multicolumn{4}{c|}{\textbf{14 Tasks}}
& \multicolumn{4}{c}{\textbf{20 Tasks}} \\
\cline{2-5}
\cline{6-9}
\cline{10-13}
& \textbf{ACC$\uparrow$} & \textbf{Gen.$\uparrow$} & \textbf{H-score$\uparrow$} & \textbf{BWT$\uparrow$}
& \textbf{ACC$\uparrow$} & \textbf{Gen.$\uparrow$} & \textbf{H-score$\uparrow$} & \textbf{BWT$\uparrow$}
& \textbf{ACC$\uparrow$} & \textbf{Gen.$\uparrow$} & \textbf{H-score$\uparrow$} & \textbf{BWT$\uparrow$} \\
\hline

Pre-Trained
& 55.4 & 66.4 & 60.4 & --
& 62.0 & 66.4 & 64.1 & --
& 59.8 & 66.4 & 62.9 & -- \\

Individual
& 92.4 & 57.3 & 70.7 & --
& 91.3 & 54.9 & 68.5 & --
& 91.6 & 53.2 & 67.3 & -- \\

\hline

Task Arithmetic
& \pmv{77.1}{0.0} & \pmv{59.9}{0.0} & \pmv{67.4}{0.0} & \pmv{-4.2}{1.0}
& \pmv{70.9}{0.0} & \pmv{65.7}{0.0} & \pmv{68.2}{0.0} & \pmv{-1.3}{0.4}
& \pmv{64.2}{0.0} & \pmv{\underline{62.0}}{0.0} & \pmv{63.1}{0.0} & \pmv{-3.6}{0.4} \\

Ties-Merging
& \pmv{66.8}{3.7} & \pmv{39.4}{9.3} & \pmv{47.8}{9.7} & \pmv{-5.5}{0.4}
& \pmv{70.3}{0.7} & \pmv{\textbf{66.1}}{0.6} & \pmv{68.1}{0.6} & \pmv{\textbf{1.4}}{0.7}
& \pmv{63.0}{1.6} & \pmv{59.7}{1.7} & \pmv{61.5}{1.5} & \pmv{\textbf{-1.5}}{1.2} \\

WUDI-Merging
& \pmv{81.0}{4.7} & \pmv{59.3}{4.0} & \pmv{68.4}{4.3} & \pmv{-12.6}{5.4}
& \pmv{75.0}{4.1} & \pmv{58.1}{2.9} & \pmv{65.5}{3.3} & \pmv{-16.9}{4.4}
& \pmv{69.6}{4.7} & \pmv{55.6}{4.5} & \pmv{61.8}{4.5} & \pmv{-18.5}{14.2} \\

Iso-C
& \pmv{78.5}{1.2} & \pmv{\underline{66.3}}{0.2} & \pmv{72.1}{0.5} & \pmv{-6.7}{0.5}
& \pmv{79.7}{1.3} & \pmv{63.7}{0.6} & \pmv{\underline{70.8}}{0.7} & \pmv{-7.1}{1.1}
& \pmv{73.0}{1.1} & \pmv{60.0}{0.6} & \pmv{65.7}{0.7} & \pmv{-9.9}{1.7} \\

KnOTS-TIES
& \pmv{57.9}{8.4} & \pmv{35.3}{7.6} & \pmv{44.9}{8.4} & \pmv{-13.5}{5.5}
& \pmv{71.6}{0.3} & \pmv{\underline{65.9}}{0.2} & \pmv{68.6}{0.2} & \pmv{\underline{1.0}}{0.1}
& \pmv{62.7}{1.1} & \pmv{57.6}{1.4} & \pmv{60.1}{1.2} & \pmv{\underline{-2.3}}{0.6} \\

TSV-M
& \pmv{75.4}{4.0} & \pmv{62.8}{2.0} & \pmv{68.4}{2.3} & \pmv{-18.5}{4.6}
& \pmv{69.2}{3.1} & \pmv{62.0}{1.7} & \pmv{65.6}{2.2} & \pmv{-22.7}{3.1}
& \pmv{63.1}{1.3} & \pmv{59.7}{3.4} & \pmv{61.2}{2.6} & \pmv{-28.2}{1.8} \\

OPCM
& \pmv{81.8}{0.3} & \pmv{64.4}{0.3} & \pmv{72.0}{0.3} & \pmv{-4.8}{0.7}
& \pmv{77.1}{0.5} & \pmv{63.8}{0.2} & \pmv{69.9}{0.2} & \pmv{-5.1}{1.4}
& \pmv{70.3}{0.2} & \pmv{\textbf{62.7}}{0.4} & \pmv{\underline{66.2}}{0.4} & \pmv{-6.3}{2.2} \\

NUFILT
& \pmv{\textbf{87.3}}{0.1} & \pmv{65.1}{0.1} & \pmv{\underline{74.6}}{0.1} & \pmv{\textbf{-1.6}}{0.5}
& \pmv{\underline{83.0}}{0.2} & \pmv{60.0}{0.2} & \pmv{69.7}{0.2} & \pmv{-3.5}{0.6}
& \pmv{\underline{78.1}}{0.9} & \pmv{54.6}{0.5} & \pmv{64.2}{0.5} & \pmv{-7.1}{1.9} \\

\textbf{\methodshort{} (Ours)}
& \pmv{\underline{86.2}}{0.3} & \pmv{\textbf{66.6}}{0.2} & \pmv{\textbf{75.1}}{0.2} & \pmv{\underline{-3.0}}{0.5}
& \pmv{\textbf{83.3}}{0.3} & \pmv{64.7}{0.1} & \pmv{\textbf{72.8}}{0.2} & \pmv{-4.3}{0.5}
& \pmv{\textbf{78.4}}{0.6} & \pmv{\underline{62.0}}{0.4} & \pmv{\textbf{69.2}}{0.4} & \pmv{-8.3}{0.9} \\

\hline
\multicolumn{13}{c}{\textbf{ViT-L/14}} \\
\hline
\multicolumn{1}{l|}{\textbf{Method}}
& \multicolumn{4}{c|}{\textbf{8 Tasks}}
& \multicolumn{4}{c|}{\textbf{14 Tasks}}
& \multicolumn{4}{c}{\textbf{20 Tasks}} \\
\cline{2-5}
\cline{6-9}
\cline{10-13}
& \textbf{ACC$\uparrow$} & \textbf{Gen.$\uparrow$} & \textbf{H-score$\uparrow$} & \textbf{BWT$\uparrow$}
& \textbf{ACC$\uparrow$} & \textbf{Gen.$\uparrow$} & \textbf{H-score$\uparrow$} & \textbf{BWT$\uparrow$}
& \textbf{ACC$\uparrow$} & \textbf{Gen.$\uparrow$} & \textbf{H-score$\uparrow$} & \textbf{BWT$\uparrow$} \\
\hline

Pre-Trained
& 64.9 & 76.7 & 70.3 & --
& 69.1 & 76.7 & 72.7 & --
& 65.6 & 76.7 & 70.7 & -- \\

Individual
& 94.3 & 68.8 & 79.5 & --
& 93.4 & 69.7 & 79.8 & --
& 93.5 & 69.4 & 79.6 & -- \\

\hline

Task Arithmetic
& \pmv{80.5}{0.0} & \pmv{70.0}{0.0} & \pmv{74.9}{0.0} & \pmv{-6.8}{1.0}
& \pmv{77.9}{0.0} & \pmv{76.1}{0.0} & \pmv{77.0}{0.0} & \pmv{-1.8}{0.3}
& \pmv{70.3}{0.0} & \pmv{73.6}{0.0} & \pmv{71.9}{0.0} & \pmv{-3.3}{0.3} \\

Ties-Merging
& \pmv{64.3}{7.0} & \pmv{52.0}{7.0} & \pmv{57.1}{7.3} & \pmv{-13.0}{5.7}
& \pmv{78.0}{0.6} & \pmv{\underline{76.4}}{0.2} & \pmv{77.1}{0.4} & \pmv{\underline{-1.1}}{0.4}
& \pmv{68.3}{0.9} & \pmv{70.8}{0.9} & \pmv{69.5}{0.8} & \pmv{\underline{-2.9}}{1.0} \\

WUDI-Merging
& \pmv{87.5}{3.3} & \pmv{71.7}{2.9} & \pmv{78.8}{3.0} & \pmv{-7.3}{3.7}
& \pmv{84.2}{3.7} & \pmv{71.5}{2.8} & \pmv{77.3}{3.1} & \pmv{-9.4}{4.0}
& \pmv{\underline{78.1}}{2.8} & \pmv{70.9}{1.6} & \pmv{74.3}{2.0} & \pmv{-15.8}{2.9} \\

Iso-C
& \pmv{86.9}{0.5} & \pmv{\underline{77.1}}{0.2} & \pmv{81.8}{0.4} & \pmv{-3.7}{0.6}
& \pmv{86.9}{1.8} & \pmv{76.1}{0.5} & \pmv{\underline{81.2}}{0.8} & \pmv{-3.7}{1.7}
& \pmv{80.9}{0.8} & \pmv{\underline{74.7}}{0.2} & \pmv{\underline{77.6}}{0.4} & \pmv{-5.5}{1.3} \\

KnOTS-TIES
& \pmv{68.3}{5.7} & \pmv{52.4}{5.6} & \pmv{59.3}{5.6} & \pmv{-11.6}{3.6}
& \pmv{78.8}{0.3} & \pmv{\underline{76.4}}{0.1} & \pmv{77.6}{0.1} & \pmv{\textbf{0.3}}{0.3}
& \pmv{69.7}{0.8} & \pmv{71.3}{0.7} & \pmv{70.6}{0.6} & \pmv{\textbf{-2.3}}{0.7} \\

TSV-M
& \pmv{82.2}{3.6} & \pmv{73.8}{1.5} & \pmv{77.7}{2.0} & \pmv{-13.0}{4.0}
& \pmv{78.1}{3.6} & \pmv{73.7}{1.7} & \pmv{75.8}{2.3} & \pmv{-15.6}{3.9}
& \pmv{70.5}{1.2} & \pmv{73.1}{0.9} & \pmv{71.8}{1.0} & \pmv{-23.3}{1.3} \\

OPCM
& \pmv{87.0}{0.4} & \pmv{75.1}{0.2} & \pmv{80.6}{0.2} & \pmv{-2.6}{1.0}
& \pmv{83.5}{0.2} & \pmv{75.1}{0.2} & \pmv{79.1}{0.2} & \pmv{-4.3}{0.7}
& \pmv{76.0}{0.2} & \pmv{\underline{74.7}}{0.2} & \pmv{75.4}{0.2} & \pmv{-6.5}{1.8} \\

NUFILT
& \pmv{\textbf{91.6}}{0.1} & \pmv{75.6}{0.1} & \pmv{\underline{82.8}}{0.1} & \pmv{\textbf{-1.1}}{0.3}
& \pmv{\underline{89.2}}{0.1} & \pmv{73.2}{0.1} & \pmv{80.3}{0.1} & \pmv{-2.0}{0.3}
& \pmv{\textbf{84.7}}{0.8} & \pmv{69.9}{0.1} & \pmv{76.5}{0.2} & \pmv{-4.6}{0.7} \\

\textbf{\methodshort{} (Ours)}
& \pmv{\underline{91.2}}{0.1} & \pmv{\textbf{77.2}}{0.1} & \pmv{\textbf{83.6}}{0.1} & \pmv{\underline{-1.8}}{0.3}
& \pmv{\textbf{89.3}}{0.2} & \pmv{\textbf{76.5}}{0.1} & \pmv{\textbf{82.4}}{0.1} & \pmv{-2.2}{0.3}
& \pmv{\textbf{84.7}}{0.4} & \pmv{\textbf{75.0}}{0.0} & \pmv{\textbf{79.6}}{0.2} & \pmv{-6.3}{0.4} \\

\bottomrule
\end{tabular}%
}
\endgroup
\label{tab:appendix_b16_l14_results}
\end{table*}

\begin{table*}[!htbp]
\centering
\caption{General performance on held-out control datasets (ImageNet, ImageNet-R, and ObjectNet) after continually merging 8, 14, and 20 tasks on ViT-B/32. Gen. Avg. is the average of the three datasets. Among merging methods, best results are in \textbf{bold} and second-best are \underline{underlined}.}
\begingroup
\setlength{\tabcolsep}{3.5pt}
\renewcommand\arraystretch{1.08}
\newcommand{\pmv}[2]{#1{\scriptsize\textsubscript{\ensuremath{\pm#2}}}}
\resizebox{\textwidth}{!}{%
\begin{tabular}{l|cccc|cccc|cccc}
\toprule
\multirow{2}{*}{\textbf{Method}}
& \multicolumn{4}{c|}{\textbf{8 Tasks}}
& \multicolumn{4}{c|}{\textbf{14 Tasks}}
& \multicolumn{4}{c}{\textbf{20 Tasks}} \\
\cmidrule(lr){2-5}
\cmidrule(lr){6-9}
\cmidrule(lr){10-13}
& \textbf{IN} & \textbf{IN-R} & \textbf{ObjNet} & \textbf{Gen. Avg.}
& \textbf{IN} & \textbf{IN-R} & \textbf{ObjNet} & \textbf{Gen. Avg.}
& \textbf{IN} & \textbf{IN-R} & \textbf{ObjNet} & \textbf{Gen. Avg.} \\
\midrule

Pre-Trained
& 62.0 & 69.3 & 45.1 & 58.8
& 62.0 & 69.3 & 45.1 & 58.8
& 62.0 & 69.3 & 45.1 & 58.8 \\

Individual
& 44.9 & 54.3 & 28.4 & 42.5
& 46.8 & 56.2 & 30.4 & 44.4
& 46.7 & 56.1 & 30.3 & 44.4 \\

\midrule

Task Arithmetic
& \pmv{45.9}{0.0} & \pmv{57.7}{0.0} & \pmv{29.8}{0.0} & \pmv{44.5}{0.0}
& \pmv{60.1}{0.0} & \pmv{69.9}{0.0} & \pmv{41.9}{0.0} & \pmv{57.3}{0.0}
& \pmv{\underline{56.3}}{0.0} & \pmv{\textbf{67.4}}{0.0} & \pmv{\underline{37.9}}{0.0} & \pmv{\underline{53.8}}{0.0} \\

Ties-Merging
& \pmv{19.3}{8.4} & \pmv{30.1}{10.3} & \pmv{14.6}{4.3} & \pmv{21.3}{7.7}
& \pmv{\textbf{61.0}}{0.2} & \pmv{\textbf{70.3}}{0.2} & \pmv{42.1}{0.4} & \pmv{\underline{57.8}}{0.3}
& \pmv{53.6}{1.5} & \pmv{65.1}{1.1} & \pmv{35.5}{1.5} & \pmv{51.4}{1.3} \\

WUDI-Merging
& \pmv{45.4}{8.0} & \pmv{54.0}{7.7} & \pmv{31.1}{6.4} & \pmv{43.5}{7.3}
& \pmv{45.4}{6.4} & \pmv{55.0}{5.9} & \pmv{30.7}{5.1} & \pmv{43.7}{5.7}
& \pmv{46.5}{3.8} & \pmv{56.6}{4.0} & \pmv{31.9}{4.8} & \pmv{45.0}{4.1} \\

Iso-C
& \pmv{\underline{58.9}}{0.7} & \pmv{\textbf{67.3}}{0.4} & \pmv{\underline{40.3}}{0.9} & \pmv{\underline{55.5}}{0.6}
& \pmv{56.1}{0.5} & \pmv{65.8}{0.6} & \pmv{38.2}{0.8} & \pmv{53.4}{0.6}
& \pmv{51.1}{0.8} & \pmv{60.7}{1.0} & \pmv{34.7}{1.2} & \pmv{48.8}{0.9} \\

KnOTS-TIES
& \pmv{20.1}{7.0} & \pmv{31.4}{7.8} & \pmv{15.9}{3.5} & \pmv{22.5}{6.1}
& \pmv{\underline{60.8}}{0.1} & \pmv{\underline{70.1}}{0.1} & \pmv{\textbf{43.0}}{0.1} & \pmv{\textbf{58.0}}{0.1}
& \pmv{53.8}{1.9} & \pmv{65.0}{1.4} & \pmv{35.8}{1.7} & \pmv{51.5}{1.7} \\

TSV-M
& \pmv{52.6}{4.6} & \pmv{61.6}{3.8} & \pmv{35.8}{4.4} & \pmv{50.0}{4.2}
& \pmv{52.9}{4.5} & \pmv{62.1}{4.1} & \pmv{35.8}{4.1} & \pmv{50.3}{4.1}
& \pmv{54.3}{3.9} & \pmv{63.6}{2.9} & \pmv{36.7}{4.8} & \pmv{51.5}{3.8} \\

OPCM
& \pmv{54.6}{0.6} & \pmv{64.2}{0.3} & \pmv{36.6}{0.4} & \pmv{51.8}{0.4}
& \pmv{56.9}{0.2} & \pmv{66.9}{0.2} & \pmv{38.3}{0.3} & \pmv{54.0}{0.1}
& \pmv{55.8}{0.4} & \pmv{\underline{66.3}}{0.2} & \pmv{37.4}{0.4} & \pmv{53.1}{0.3} \\

NUFILT
& \pmv{55.5}{0.3} & \pmv{62.6}{0.4} & \pmv{38.6}{0.2} & \pmv{52.3}{0.3}
& \pmv{52.0}{0.2} & \pmv{60.6}{0.4} & \pmv{35.7}{0.4} & \pmv{49.4}{0.2}
& \pmv{46.4}{0.5} & \pmv{55.5}{0.5} & \pmv{31.7}{0.3} & \pmv{44.5}{0.4} \\

\methodshort{} (Ours)
& \pmv{\textbf{60.0}}{0.2} & \pmv{\underline{67.2}}{0.3} & \pmv{\textbf{42.9}}{0.2} & \pmv{\textbf{56.7}}{0.2}
& \pmv{59.0}{0.2} & \pmv{67.4}{0.2} & \pmv{\underline{42.7}}{0.3} & \pmv{56.4}{0.2}
& \pmv{\textbf{57.4}}{0.5} & \pmv{65.8}{0.3} & \pmv{\textbf{41.6}}{0.3} & \pmv{\textbf{54.9}}{0.4} \\

\bottomrule
\end{tabular}%
}
\endgroup
\label{tab:general_performance}
\end{table*}

\section{Discussions}
\label{appdix:Discussions}
\subsection{Limitations}
\label{appdix:Limitations}
\methodshort{} assumes that all checkpoints share the same pretrained initialization and architecture, enabling comparable task vectors and risk estimates. Extending it to mismatched initializations or heterogeneous backbones requires cross-model parameter or subspace alignment, which remains future work.

\subsection{Broader Impacts}
\label{appdix:Broader Impacts}
This paper aims to advance the Machine Learning field. Our work has potential societal impacts, but none require specific highlighting here.

\subsection{LLM Usage}
In preparing this submission, we used large language models (LLMs) solely as an assistive tool for sentence-level editing, including grammar correction, spelling adjustments, and minor word-choice refinements. The LLM was not involved in research ideation, methodological design, experimental analysis, or content generation beyond language editing. All substantive scientific contributions are solely those of the authors.

\end{document}